\documentclass{article}
\usepackage{amsmath}%
\usepackage{amsfonts}%
\usepackage{amssymb}%
\usepackage{amsthm}%
\usepackage{appendix}%
\usepackage{theoremref}%
\usepackage{graphicx}%
\newcommand{\BE}{{\mathbb E}}%
\newcommand{\BR}{{\mathbb R}}%
\newcommand{\BZ}{{\mathbb Z}}%
\newcommand{\calI}{{\mathcal I}}%
\newcommand{\calM}{{\mathcal M}}%
\newcommand{\calP}{{\mathcal P}}%
\newcommand{\CR}{\par\noindent}
\newcommand{\smallspace}{\vskip 10pt \noindent}
\newcommand{\arrow}{\rightarrow}
\newcommand{\eps}{\epsilon}
\newcommand{\QEDA}{\hfill\ensuremath{\blacksquare}}
\newtheorem{theorem}{Theorem}
\newtheorem{lemma}{Lemma}
\newtheorem{corollary}[lemma]{Corollary}

\newcommand{\thistheoremname}{}
\newtheorem*{genericthm*}{\thistheoremname}
\newenvironment{namedthm}[1]%
  {\renewcommand{\thistheoremname}{#1}%
   \begin{genericthm*}}%
  {\end{genericthm*}}

\begin{document}

\title{Exact marginal inference in \\ Latent Dirichlet Allocation}
\author{ Hartmut Maennel\\ {\small Google Research, Brain Team, Z\"urich} } 
\date{}
\maketitle

\begin{abstract}
\noindent
Assume we have potential ``causes'' $z\in Z$, which produce ``events'' $w$
with known probabilities $\beta(w|z)$. 
We observe $w_1,w_2,...,w_n$, what can we say about the distribution of the causes? 
A Bayesian estimate will assume a prior on distributions on $Z$ 
(we assume a Dirichlet prior) and calculate a posterior.
An average over that posterior then gives a distribution on $Z$, which estimates how much each cause
$z$ contributed to our observations.\CR
This is the setting of Latent Dirichlet Allocation, which can be applied e.g. to topics ``producing'' words 
in a document.
In this setting usually the number of observed words is large, but the number of potential topics is small.
We are here interested in applications with many potential ``causes'' (e.g. locations on the globe),
but only a few observations.\CR
We show that the exact Bayesian estimate can be computed in linear time (and constant space) in $|Z|$ 
for a given upper bound on $n$ with a surprisingly simple formula.
We generalize this algorithm to the case of sparse probabilities $\beta(w|z)$, in which we only need to assume
that the tree width of an ``interaction graph'' on the observations is limited. \CR
On the other hand we also show that without such limitation the problem is NP-hard.
\end{abstract}

\section{Problem description}
Assume we have a fixed set of ``causes'' $z\in Z$, which produce ``events'' $w$ 
from another fixed set $W$ with known probabilities $\beta(w|z)$. 
We observe $w_1,w_2,...,w_n$, and we assume that these events correspond to (unknown) causes $z_1, z_2,...,z_n$.
(The same cause can appear several times in the sequence $z_1,...,z_n$.)
Given the observations and $\beta$, what can we infer about the distribution of causes that produced them?
\smallspace
Of course, usually we cannot reconstruct the individual causes $z_i$ that produced the observations $w_i$, 
instead we are looking for a probability distribution $\theta$ on the set $Z$ which describes 
the ``mixture'' of the causes that produced our observations, i.e. $\theta(z)$ can be interpreted as the 
probability that the next observation would come from the cause $z$. 
Can we determine/estimate $\theta$ from the  observations?
\smallspace
If we have ``infinitely many'' observations that give us exact probabilities
\begin{equation}
    p(w) = \sum_{z\in Z} \beta(w|z)\cdot \theta(z)
    \label{eq:linear}
\end{equation}
this just means solving the system of linear equations \eqref{eq:linear}. For finitely many observations, we 
can ask for the maximum likelihood solution $\theta$ with
\begin{equation}
     \prod_{i=1}^n \sum_{z\in Z} \beta(w_i|z)\cdot \theta(z) \arrow \max
     \label{eq:maxLikelihood}
\end{equation}
which we can compute using the Expectation Maximization algorithm. \CR
However, in particular for smaller number of observations, the maximum likelihood solution can be
very misleading since there may be different $\theta$ 
which give almost the same likelihood \eqref{eq:maxLikelihood}. For example, it may assign
probability zero to some causes which are likely to have a large effect, 
see appendix B1 for a simple example.
To avoid this, the Latent Dirichlet Allocation setting will use a Bayesian approach for estimating $\theta$.
\smallspace
One application in which this basic problem occurs is topic classification for texts. 
The ``aspect'' or ``topic mixture'' model for texts (\cite{Hofmann}, \cite{BleiNgJordan}) 
states that a text is about a mixture of topics, each word in the text ``is caused by'' one of these topics, 
and for each topic $z$ there are fixed probabilities $\beta(w|z)$ that a word $w$ appears because of the topic $z$.
For long documents, it is not difficult to find the topic mixture, but for very short texts 
(e.g. tweets, book/paper titles, short descriptions of photos/videos) there may be a considerable uncertainty.
\CR
Due to the fundamental nature of this problem, it occurs in many different application areas. In fact, in the probably
earliest description of the Bayesian approach to this problem \cite{Pritchard}, 
the ``causes'' were populations (e.g. of a species of birds originating at a particular location),  
and the ``observations'' were genetic markers (microsatellites). Another example where $Z$ would 
be a (very large) set of locations is determining the locations of a group of photos from clues in the images.
\smallspace
Returning to the Bayesian problem statement in Latent Dirichlet Allocation, we specify as additional input
a prior for $\theta$. The usual choice for a prior in this situation is a Dirichlet distribution, 
it allows us to specify both how likely we think the causes $z$ are initially and how much weight 
we want to put on this initial assumption compared to the observations.\CR
So in this Bayesian formulation, our initial problem turns into the following concrete computational problem:
\CR
Given the vector $\alpha= (\alpha_1,...,\alpha_m)$ with $m=|Z|$ which determines the prior $Dir(\alpha_1,...,\alpha_m)$
for the mixture $\theta$ and the matrix $\beta=(\beta(w|z))$,
which determines the probability that event $w$ is produced by cause $z$, 
we get as the probability to observe $w_1,...,w_n$ the expression\CR
\begin{eqnarray}
  p(w_1,...,w_n | \alpha, \beta)
 = \int_{\theta\in\Delta} \frac{\theta^{\alpha-1}}{B(\alpha)}
     \prod_{i=1}^n \sum_{z \in Z} \beta(w_i | z) \cdot \theta(z)
    \ d\theta  
    \label{eq:def_p}
\end{eqnarray}
where $\Delta$ is the $(m-1)$--dimensional simplex in $\BR^m$ given by 
$0 \leq \theta_1,\theta_2,...,\theta_m \leq 1$ and $\theta_1 +\theta_2 +... +\theta_m = 1$, 
and we use the abbreviations  $  |\alpha| := \sum_{j=1}^m \alpha_j $ and 
\[
   \theta^{\alpha - 1} := \prod_{j=1}^m \theta_j^{\alpha_j-1}
   \ \hbox{,}\ 
   \Gamma(\alpha) := \prod_{j=1}^m \Gamma(\alpha_j)
      \ \hbox{,}\ 
   B(\alpha) := \frac{\Gamma(\alpha)}{\Gamma(|\alpha|)}
             = \frac{\prod_{j=1}^m \Gamma(\alpha_j)}{\Gamma(\sum_{j=1}^m \alpha_j)}  
\]
This ``generative model'' is conveniently summarized in this diagram:
\smallspace
\includegraphics[width=0.7\textwidth]{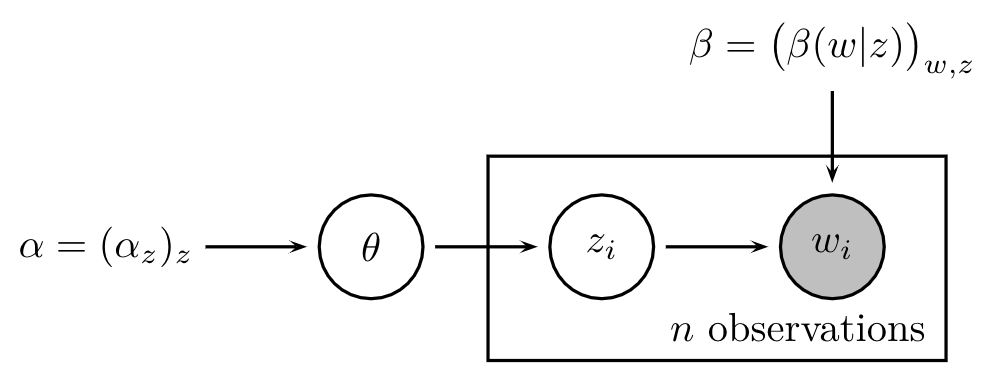}
\smallspace
For the  Bayesian estimate for the mixture component $\theta(z)$ we get
\begin{eqnarray}
     \BE[\theta(z)|w_1,...,w_n]\ =\  
     \frac {
     \int_{\theta\in\Delta} \theta(z)\cdot \frac{\theta^{\alpha-1}}{B(\alpha)}
    \prod_{i=1}^n \sum_{z' \in Z} \beta(w_i | z') \cdot \theta(z')
    \ d\theta}
    { p(w_1,...,w_n|\alpha,\beta)}  \label{eq:BayesEst}
\end{eqnarray}
So the question we will investigate is: How difficult is it to compute \eqref{eq:BayesEst} 
from the input $\alpha,\beta$, and the observations $w_1,...,w_n$?
\smallspace
These expressions are difficult to evaluate exactly, the na\"ive evaluation even of the simple
\eqref{eq:def_p} would give $O(m^n)$ terms. \CR
The standard solutions use approximations - either variational approximations or Gibbs sampling. 
Sampling is a stochastic approximation and it may be difficult to 
estimate the error. The variational approximation given in \cite{BleiNgJordan}
is deterministic and fast, but can give
substantially different results from the exact solution if we have few events and a low prior 
(low numbers $\alpha(z)$ are common for large sets of potential causes since their sum 
gives the total amount of evidence / weight assigned to the prior).
Below is a toy example with 3 causes and 2 observations:
\CR
\vskip 3mm
\begin{tabular}{r||l|l|l}
  &$z_1$ & $z_2$  & $z_3$ \\
  \hline
  $\beta(w_1|z)$& 0.09 & 0.05 & 0.02 \\
  $\beta(w_2|z)$& 0.02 & 0.05 & 0.08 \\
  $\alpha(z)$ & 1/3 & 1/3 & 1/3 \\
  \hline 
  Max. likelihood & 0.524 & 0 & 0.476 \\
  Variational Bayes & 0.446 & 0.151 & 0.403 \\
  Exact Bayes &  0.331 & 0.355 & 0.314   \\
\end{tabular}
\CR\vskip 2mm
(See appendix B for details.) 
\smallspace
So it is interesting to ask whether it is possible to compute the exact value of \eqref{eq:BayesEst} in 
any case of interest.

\section{Our contributions}
We are investigating the complexity of evaluating \eqref{eq:BayesEst} exactly. Since the Bayesian formulation of our initial question is interesting in particular for small $n$, we will mainly investigate the complexity of this problem for fixed (small) $n$.

We will see that this question has basically the same answer for the probability \eqref{eq:def_p} 
and the expectation \eqref{eq:BayesEst}. A na\"ive computation would already need $O(m^n)$ for \eqref{eq:def_p}.

{\bf On the positive side}, we give a surprisingly simple expression for \eqref{eq:def_p}, see  equation \eqref{eq:SimpleFormula}. 
For fixed (small) $n$ 
this can be evaluated in time $O(m)$ and with space requirements $O(1)$, the same is true for computing
\eqref{eq:BayesEst} with equation \eqref{eq:BayesianEstimate}, see Theorem 1 and 1'.
\CR
These expressions \eqref{eq:SimpleFormula}, \eqref{eq:BayesianEstimate} are even useful for infinite spaces of
``causes'' $z$ for which the probabilities $\beta(w|z)$
are given as functions of $z$ and $\alpha$ is a measure on $Z$ which defines a Dirichlet process prior, see section \ref{sec:TopicMixtures}.
\CR
These expressions can be evaluated in a way that requires $O(2^n\cdot m + 3^n)$ operations (Theorem 1 and 1'), 
this is perfectly feasible for e.g. 15 observations and
a large number of possible causes. On the other hand, it does not seem feasible for e.g. 30 observations.
If the probabilities $\beta(w|z)$ are sparse (i.e. many are 0 or too small to matter), this can be relaxed:
Define an undirected graph with the observations as nodes, and connect two observations $w_i, w_j$ if
there is a cause that can explain both (i.e. $p(w_i|z) > 0,\ p(w_j|z) > 0$). Then the exponential
dependence is not on $n$, but only on the tree width of the resulting graph (Theorems 4, 4'). 
\CR
These results can be used to compute \eqref{eq:def_p} and \eqref{eq:BayesEst} exactly in the case of
small $n$ or 
small tree width, and may also open new possibilities for approximations for larger $n$.
\smallspace
{\bf On the negative side}, we show with a dimension argument that the factor $2^n$ cannot be decreased for any 
algorithm that has the same general structure of our proposed method, i.e. makes one pass over the causes (Theorem 2, 2'). Even without
restriction on the structure of an exact calculation, it seems unlikely that a polynomial running time could 
be achieved in general: We show this would imply we could also compute the permanent of a 0-1 matrix in
polynomial time, which is known to imply NP=P (Theorem 3, 3').

\section{Related work}
In their article \cite{BleiNgJordan} introducing Latent Dirichlet Allocation, the authors use this generative topic mixture model for words in documents as a building block for their LDA method (which deals with a more complicated problem than our fundamental problem, since it does not assume the topics and probabilities are given). They remark that \eqref{eq:def_p} ``is intractable due to the coupling between $\theta$ and $\beta$ in the summation over latent topics'' and quote \cite{Dickey1}, which gives ways to compute such expressions. While this gives an indication that the occurring expressions are difficult to evaluate, it does not formally prove that there cannot be other efficient algorithms.
\smallspace
This question was more thoroughly investigated in \cite{SontagRoy}. 
Section 4 in that paper treats Marginal Inference, however the authors do not consider the question of exact computations, but treat
the more common case of approximation by sampling. They use a constant $\alpha(z)=\alpha$ and determine the difficulty in two regimes:
\begin{itemize}
  \item[a)] $\alpha > 1$: In that case approximate results via sampling can be obtained in polynomial time.
     However, they emphasize that this is a theoretical result (they mention a constant of $10^{30}$).
  \item[b)] For ``extremely small'' $\alpha$ even approximation becomes NP hard.
\end{itemize}

In contrast, in this paper we investigate exact solutions. This makes a difference also in the negative results:
We show that an efficient exact algorithm for \eqref{eq:def_p} would allow computing the 
permanent of a 0-1 matrix efficiently, 
this is known to be a \#P hard problem. But  it is also known that the permanent
can be approximated in polynomial time (see e.g. section 17.3.2 in \cite{AroraBarak}).

\section{Factorizing the integration}
In the following we will first focus on the computation of \eqref{eq:def_p} [the probability to observe $w_1,...,w_n$] and then extend the results to \eqref{eq:BayesEst} [the Bayesian estimate for the mixture component] in section \ref{sec:TopicMixtures}.

To sample from a Dirichlet distribution $Dir(\alpha_1,...,\alpha_m)$, a common method is to generate 
$m$ independent samples from the Gamma distributions $\Gamma(\alpha_i, 1)$ and divide by their sum.
This essentially is the case $h=0$ of the following Lemma.\CR
\begin{lemma}\thlabel{DirichletGamma}
Let $f:\BR_{>0}^m\arrow \BR$ be a function that is homogeneous of degree $h$, i.e. 
\[
   f(t\cdot \theta) = t^h\cdot f(\theta)
   \qquad \hbox{for} \quad
   t > 0
\]
then 
\[
   \BE_{\theta\sim Dir(\alpha,1)}[f(\theta)] = 
   \frac{\Gamma(|\alpha|)}{\Gamma(|\alpha|+h)} \BE_{\theta\sim \Gamma(\alpha,1)}[f(\theta)]
\]
where $\theta\sim\Gamma(\alpha,1)$ means that the $\theta_j$ follow $m$ independent Gamma distributions $\Gamma(\alpha_j,1)$.\CR
\end{lemma}
\CR
(Proof in appendix C.)
\CR
Applying this to the special case $h:=n$ and 
\[
   f(\theta) := \prod_{i=1}^n \sum_{z \in Z} \beta(w_i | z) \cdot \theta(z)
\]
gives the following Corollary.\CR
\begin{corollary}
\begin{eqnarray*}
\lefteqn{p(w_1,...,w_n | \alpha, \beta)} \\
   &=& \frac{\Gamma(|\alpha|)}{\Gamma(|\alpha| + n)} \cdot
    \BE_{\theta\sim\Gamma(\alpha,1)} \left[
    \prod_{i=1}^n \sum_{z \in Z} \beta(w_i | z) \cdot \theta(z)
    \right]
\end{eqnarray*}
\end{corollary}
\CR
In this expression the expectation / integral is factorized into $m$ one-dimensional integrals 
corresponding to the causes in $Z$, but the function does not 
factorize over the causes, so we cannot write it as a product of expectations. (The main point
in the next section is that this changes when we look instead at the generating function.)
\CR
We abbreviate this expectation (or ``unnormalized probability'') by $\tilde p$ and rewrite it as
\begin{eqnarray}
 \tilde{p}(w_1,...,w_n | \alpha, \beta)
  &:=& \BE_{\theta\sim\Gamma(\alpha,1)} \left[\prod_{i=1}^n \sum_{z \in Z} \beta(w_i | z) \cdot \theta(z)\right] \label{eq:def_p_tilde} \\
  &=& \sum_{\vec{z}\in Z^n}  \prod_{i=1}^n \beta(w_i | z_i) \cdot \BE_{\theta\sim\Gamma(\alpha,1)} \left[\prod_{i=1}^n \theta(z_i)\right] \nonumber
\end{eqnarray}
Now for each $(z_1,...,z_n)$ the last expectation only contains powers of some $\theta(z)$, so this factors over the causes
and can be written as a product of moments of the one dimensional Gamma distributions. The result for $\tilde p$ will be a sum 
of different contributions for each partition of the observations $(w_1,...,w_m)$ which specifies
which of the observations $w_i$ come from the same cause in $Z$.
While one could do computations along these lines for the first few $n=1,2,3,...$, the calculation looks complicated in general.
\smallspace
However, we will see that in fact the end result is very simple, and can be obtained elegantly using the combinatorial tool of 
generating functions.

\section{Generating functions}
The easiest way to list all moments of the (one dimensional) Gamma distribution is to describe the moment generating 
function (e.g. \cite{Wasserman}, chapter 3.6):
\begin{equation}
   \BE_{\theta\sim \Gamma(\alpha,1)}\left[ e^{\theta X}\right] = 
   \sum_{i=0}^\infty  \BE_{\theta\sim \Gamma(\alpha,1)}\left[ \theta^i\right]\cdot \frac{X^i}{i!}
   = (1 - X)^{-\alpha}  \label{eq:moments}
\end{equation}
Here a sequence of numbers $c_i = \BE[\theta^i]$ (the moments) is ``encoded'' in one function by considering
the power series $\sum_{i=0}^\infty c_i \cdot X^i/i!$. This is a common tool in combinatorics --- often sequences
of numbers can be described and manipulated efficiently in the form of these ``generating functions''.\CR
This function can be considered either as a formal power series, or as an analytic function of one variable (if
it converges in a neighborhood of $X=0$). In either case, knowing the function is equivalent to knowing all
the numbers $c_i$.\CR
If the numbers we are interested in depend not on one natural number $i$, but on several numbers $i_1,...,i_n$,
we can consider (formal or analytic) power series in $n$ variables. This is what we will do here --- it turns out
to be beneficial to not consider one specific probability $p(w_1,...,w_n|\alpha,\beta)$, but all these probabilities
for all finite sequences which can be formed from the observations in $W$ (where we are allowed
to repeat any observation any number of times).
\CR
To formulate such a generating function for the $\tilde p(w_1,...,w_n|\alpha,\beta)$ we introduce formal variables
$X_w$ for each possible observation $w\in W$. We use the set $\calM(W)$ of multisets of observations in $W$; 
when $I$ is the multiset which contains the observations $w_k$ with multiplicity $i_k\geq 0$, we define
\begin{eqnarray*}
   X^I :=  X_{w_1}^{i_1}\cdot X_{w_2}^{i_2}\cdot ... X_{w_n}^{i_n} &,& 
   I! := i_1!\cdot...\cdot i_n! \ ,\quad |I| := i_1 + ... + i_n \\
   \sum_{w\in I} f(w) := \sum_{k=1}^n  i_k \cdot f(w_k) &,&
   \prod_{w\in I} f(w):= \prod_{k=1}^n  f(w_k)^{i_k}
\end{eqnarray*}
The value $\tilde p(\tilde w_1,...,\tilde w_r|\alpha,\beta)$ does not change when we change the order of 
the observations $\tilde w_1,...,\tilde w_r$, so we can write 
$\tilde p(I | \alpha, \beta) := \tilde p(\tilde w_1,...,\tilde w_r| \alpha, \beta)$ if $I$ is the 
multi-sets of observations $\tilde w_1,...,\tilde w_r$. For $I=\emptyset$ we set $\tilde p(\emptyset|\alpha,\beta):= 1$.
\CR
Then we can give the generating function for the $\tilde p$ explicitly by the next Lemma.
(Specifying the generating function determines their coefficients $\tilde p$, so this is 
already a concrete formula for the $\tilde p$, but we will get a more direct formula below).
\CR
\begin{lemma}\thlabel{GenerateS}
\begin{eqnarray}
   S(X) 
   &:=& \sum_{I\in \calM(W)} \ \tilde{p}(I|\alpha,\beta) \cdot \frac{X^I}{I!} \nonumber \\
   &=&  \prod_{j=1}^m
    \Big(1-\sum_{w\in I} \beta(w|z_j)\cdot X_w\Big)^{-\alpha_j}
    \label{eq:product_power_alpha}
\end{eqnarray}
\end{lemma}
\noindent
(Proof in appendix D, the key idea is to write $S$ as the expectation of a generating function which 
factorizes over the causes $Z$;
and for one cause the result is easily derived from the moment generating function of the Gamma distribution.)

\smallspace
Now we can transform the expression \eqref{eq:product_power_alpha} to obtain a form of the generating function from which
we can read off a formula for the individual coefficients $\tilde p(I|\alpha,\beta)$. This formula is particularly simple
for the $I$ which are sets, i.e. in which all multiplicities are 0 or 1.
\CR
To write the formula concisely, denote
\[
   \beta_I(z) := \prod_{w\in I} \beta(w|z)\ , \ \ 
   \langle \beta_I \rangle := \sum_z \alpha(z) \beta_I(z)
\]
Then we can obtain from Lemma \ref{GenerateS}:\CR
\begin{lemma}\thlabel{pExplicit}
All coefficients $\tilde p(I|\alpha,\beta)$ of $X^I$ for $I$ a set 
(i.e. all multiplicities $\leq 1$) in the power series $S(X)$ 
agree with the coefficients of the polynomial
\begin{equation}
  \prod_{\emptyset \neq J\subseteq W}
        \Big(1 + \langle \beta_J \rangle\cdot \Gamma(|J|)\cdot X^J\Big)
  \label{eq:AllDiffWordsProd}
\end{equation}
Equivalently, we can write $\tilde p(I|\alpha,\beta)$ for a set $I$ as a sum over all partitions $\pi$ of $I$ into subsets:
\begin{equation}
    \tilde{p}(I|\alpha, \beta) = \sum_{\pi\in Part(I)} \prod_{J\in\pi}
    \langle \beta_J \rangle \cdot \Gamma(|J|)
   \label{eq:SimpleFormula}
\end{equation}
\end{lemma}
\noindent
(Proof in appendix E)
\smallspace
Remark 1: This formula is valid even if some of the observations are equal - in that case we have just ``wasted'' formal variables for observations that we could have labeled with the same variable. If we instead are more ``economical'' with our formal 
variables, we get an expression that is faster to calculate, but the formula is more difficult to write down without using
the formal power series.
\CR
Remark 2: It is tempting to conjecture that the contribution of partition $\pi\in Part(I)$ corresponds to the sum over all 
$\vec z$ in which the $z_j$ are equal for all $j\in J$ for $J\in\pi$. However, this is not true, these contributions
lead to more complicated terms, but most of the terms cancel to give \eqref{eq:SimpleFormula}.
\section{Complexity of the general problem}
\label{sec:complexity}
First we look at the general case, in which all $n$ observations are different and all probabilities positive, and consider the task of computing $\tilde{p}(W) = \tilde{p}(w_1,...,w_n|\alpha, \beta)$.\CR
We will see in the next section that we can save time if we know that a lot of the $\beta_J$ are 0.
\CR
\begin{theorem}
The $p(w_1,...,w_n|\alpha,\beta)$ can be computed in time $O(3^n + m\cdot 2^n)$ and space $O(2^n)$, where $m=|W|$.
\end{theorem}
\noindent
(We get this from \eqref{eq:AllDiffWordsProd}, proof in appendix F)
\CR 
Can we do even better? 
Our input $\alpha,\beta$ only has $n+n\cdot m$ numbers, so do we really need to update $O(2^n)$ values at every step?\CR
It turns out that if we keep this structure of the algorithm (i.e. one pass over the causes), we really need to keep $2^n-1$ numbers updated - this is based on a dimension argument: After processing all causes except one, we get $\tilde p(W)$ as a polynomial of degree $\leq (1,1,...,1)$ in the remaining $n$ probabilities $\beta(w_1|z_m),...,\beta(w_n|z_m)$.
These polynomials have $2^n$ coefficients, so we can consider them as subset of $\BR^{2^n}$. 
We compute the Hausdorff dimension of this set:\CR
\begin{lemma}\thlabel{Hausdorff}
Let $n, \alpha$ be fixed; then for any $m\geq 2^n$ and $\beta(w_i|z_j)$ given for $j\leq m-1$, the 
function $\BR^n\arrow\BR$ given by 
\[
    \beta(w_1|z_m),...,\beta(w_n|z_m) \mapsto \tilde p(w_1,...,w_n|\alpha, \beta)
\]
is a polynomial of degree $\leq (1,1,...,1)$.\CR
Varying the inputs $\beta(w_i|z_j)$ given for $j\leq m-1$ we obtain
a subset of the $2^n$-dimensional vector space of all possible polynomials, 
this subset has Hausdorff dimension $2^n-1$.
\end{lemma}
\noindent
(Proof in appendix G)
\CR
This allows us to prove\CR
\begin{theorem}
Any algorithm that computes $p(w_1,...,w_n|\alpha,\beta)$ exactly
\begin{itemize}
  \item using Lipschitz continuous functions and finitely many \textbf{if} statements
  \item going once through the causes $z_j$ and reading in the $\alpha(z_j), \beta(w_i,z_j)$ in order of ascending $j$
  \item and outputs after reading the data of cause $z_j$ what $p(w_1,...,w_n|\alpha,\beta)$ would be if this was the last cause,
\end{itemize}
needs space $O(2^n)$ and time $O(m\cdot 2^n)$.
\end{theorem}
\noindent
Proof: 
If there was such a one-pass algorithm that updates less than $2^n-1$ numbers at each cause, it would be possible to get a $2^n-1$ dimensional set of numbers from a smaller dimensional set of numbers in our algorithm, i.e. the last part of our algorithm could be used to describe a dimension increasing function.\CR
But a function that is built using finitely many \textbf {if} statements from Lipschitz continuous functions cannot increase the Hausdorff dimension (see e.g. \cite{Falconer}, p.32, Corollary 2.4.).
\QEDA

What happens if we drop the ``online'' requirement - could there be a polynomial time algorithm? Unfortunately, this seems unlikely, since it would imply $P=NP$:\CR
\begin{theorem}
If there is a polynomial time algorithm to compute exactly the $p(w_1,...,w_n|\alpha,\beta)$,
there is also a polynomial time algorithm to compute exactly the permanent of a 0-1 matrix, in particular this
would imply P=NP.
\end{theorem}
\noindent
Proof in appendix H.\CR
Idea: Use $n=m$ and constant $\alpha(z)=\alpha$; since $p(w_1,...,w_n|\alpha,\beta)$ is then a 
polynomial in $\alpha$, we can use the solution even at $\alpha=-1$, which is related to the permanent of the matrix $\beta$.
\CR
In fact, computing the permanent of a 0-1 matrix is known to be ``\#P-hard'', which leads to more implications
than just ``NP-hard'': E.g. even if $P\neq NP$ a subexponential time algorithm for computing the $p(w_1,...,w_n|\alpha,\beta)$
exactly would also imply subexponential time algorithms for (the permanent and) all NP problems 
(even all problems in the polynomial hierarchy), see ``Toda's theorem'' (chapter 17.4) in \cite{AroraBarak}.

\section{Sparse $\beta(w|z)$}
To simplify notation, we will assume in the following that all observations are different, however this assumption is not necessary, the same arguments work in general.
In the previous section we proved that there is no way around the exponential dependency upon the number of observations, at least if the overall structure of the algorithm remains the same, i.e. we do one pass over the causes.\CR
The exponential dependency on $n$ was ``caused'' by the fact that we have to consider
\[
   \langle \beta_J \rangle  = \sum_z \alpha(z) \prod_{w\in J} \beta(w|z)
\] 
for all $2^n$ subsets $J$ of the $n$ observations. However, depending on the area to which this is applied, most of these products may actually be zero or at least very small. \CR
We define an ``interaction graph'' between the observations: It is an undirected graph with the observations as nodes and
edges connecting any two observations $w_i, w_j$ for which 
there is a cause $z$ with $\beta(w_i|z)>0$ and $\beta(w_j|z)>0$, i.e. a cause $z$ which can generate both
observations $w_i$ and $w_j$.
\CR
(If we are interested in approximations, this could be replaced with ``$\langle \beta_{\{i,j\}}\rangle $ is small'').
\CR 
If this graph is sparsely connected, we can compute $p(w_1,...,w_n|\alpha,\beta)$ faster: E.g. if this graph is a tree,
we can compute it in $O(n\cdot m)$. More generally:\CR
\begin{theorem}
If the interaction graph has tree width $w$, and $|Z|=m$, we can compute
$p(w_1,...,w_n|\alpha,\beta)$ in time $O(n\cdot 3^w + m\cdot n\cdot 2^w)$.
\end{theorem}
\noindent
Proof in appendix I, see there or e.g. \cite{Diestel}, chapter 12.4 for the definition of tree-width.
For example, trees have the tree-width one, a
$n\times n$ grid with $n^2$ nodes has tree width $n$, and any graph with $n$ nodes has a tree width $\leq n-1$. 
So this algorithm is faster than our previous one when the tree width is significantly (at least by $\log_2(n)$) 
smaller than $n$.

\section{Bayesian estimate of the mixtures}
\label{sec:TopicMixtures}
So far we have only considered the probabilities \eqref{eq:def_p}, but in fact it is an equivalent problem to determine the expected value for the cause mixtures. To see this, note that
\begin{eqnarray*}
    \lefteqn{\BE[\theta(z)|w_1,...,w_n]} \\
     &=& \frac {
     \int_{\theta\in\Delta} \theta(z)\cdot \frac{\theta^{\alpha-1}}{B(\alpha)}
    \sum_{\vec{z}\in Z^n} \prod_{i=1}^n \beta(w_i | z_i) \cdot \theta(z_i)
    \ d\theta}
    { p(w_1,...,w_n|\alpha,\beta)} \\
    &=& \frac{p(w_z,w_1,...,w_n|\alpha,\beta)}{p(w_1,...,w_n|\alpha,\beta)}
\end{eqnarray*}
where we introduced a ``virtual observation'' $w_z$ with $\beta(w_z|z) = 1$ and $\beta(w_z|z^\prime)=0$ for $z^\prime \neq z$. This ``virtual observation'' cannot really be interpreted in terms of observations and causes, since it would increase the sum of all $\beta(w|z)$ for this $z$ to above 1. However, for the computations we made no assumption on the sum of all $\beta(w|z)$, in fact the results for $p$ and $\tilde p$ are just multiplied with a constant $c$ if we multiply all $\beta(w|z)$ by $c$ for one given $z$.\CR
This can be used to derive an explicit formula also for our expectations:\CR
\begin{lemma}\thlabel{mixtureFormula}
\begin{equation}
   \BE[\theta_z|w_1,...,w_n] = \frac{\alpha(z)}{n+|\alpha|} 
       \sum_{J \subseteq W} \beta_J(z) \cdot |J|! \cdot \frac{\tilde p(W \setminus J)}{\tilde p(W)} 
   \label{eq:BayesianEstimate}
\end{equation}
\end{lemma}
\noindent
(Proof in appendix J)
\smallspace
With this we can formulate the corresponding theorems for these expectations:\CR
\begin{namedthm}{Theorem 1'}
$\BE[\theta_z]$ can be computed in time $O(3^n + m\cdot 2^n)$ and space $O(2^n)$.
\end{namedthm}
\noindent
Proof: We compute the coefficients $\tilde p(J)/\tilde p(W)$ for all $2^n$ subsets $J\subseteq W$ and then compute \eqref{eq:BayesianEstimate} again in an online computation in time $O(m\cdot 2^n)$ for all $m$ causes - that is, the complete procedure goes twice over the set of all causes, and the  space needed is independent of the number of causes.
\QEDA
\CR
\begin{namedthm}{Theorem 4'}
If the interaction graph has tree width $w$, we can compute $\BE[\theta_z]$ in time $O(n\cdot 3^w + m\cdot n\cdot 2^w)$\CR
\end{namedthm}
\noindent
Proof: We would have $O(n\cdot 2^w)$ sets $J\subseteq W$ for which there is a nonzero $p_J(z)$,
this then also reduces the number of coefficients in \eqref{eq:BayesianEstimate} that we have to compute.
\QEDA
\CR
The equation \eqref{eq:BayesianEstimate} also has other consequences: We see that $\BE[\theta_z]$
as a function of $z$ is a linear combination of the functions $\alpha(z)\beta_J(z)$. This still makes sense for infinite $Z$, (e.g. in an idealized version of the the image location problem example) if instead of the Dirichlet distribution with prior $\alpha$ we use a Dirichlet process, the expressions $\langle \beta_J \rangle$ become integrals and then the above computation is still valid and gives that the inferred functions $\BE[\theta_z]$ are e.g. continuous / smooth / polynomial if the input functions $\alpha, \beta_i$ are in the corresponding ring of functions.
\CR
The equation \eqref{eq:BayesianEstimate} also lets us adapt the arguments of section \ref{sec:complexity}
that this procedure is optimal for an ``online'' algorithm:\CR
\begin{namedthm}{Theorem 2'}
Any algorithm that computes $\BE[\theta_z|w_1,...,w_n]$ exactly
\begin{itemize}
  \item using Lipschitz continuous functions and finitely many \textbf{if} statements
  \item going once through the causes $z_j$ and reading in the $\alpha(z_j), \beta(w_i,z_j)$ in order of ascending $j$
  \item and outputs after reading the data of cause $z_j$ what $\BE[\theta_z|w_1,...,w_n]$ would be 
  if this was the last cause,
\end{itemize}
needs space $O(2^n)$ and time $O(m\cdot 2^n)$.
\end{namedthm}
\noindent
(Proof in appendix L.)
\CR
 Note that our algorithm of Theorem 1' is ``online'' in this sense: We can output at each point what 
 $\BE[\theta_z|w_1,...,w_n]$ would be if this was the last cause,
we only do a second pass to read again the $\alpha(z), \beta_J(z)$ for all the previous causes $z$ for which 
we want to output 
$\BE[\theta_z|w_1,...,w_n]$, but we don't have to do that if we output this expectation only for the last $z$.
\smallspace
Finally, we also get a version of Theorem 3 for these expectations:
\begin{namedthm}{Theorem 3'}
If there is a polynomial time algorithm to compute exactly the $\BE[\theta_z|w_1,...,w_n]$,
there is also a polynomial time algorithm to compute exactly the permanent of a 0-1 matrix, in particular this
would imply P=NP.
\end{namedthm}
\noindent
Proof in appendix M, basic idea is to add a new virtual cause $z'$ with $\beta(w|z')=\epsilon$ for all $w$. Then
we can reconstruct $\tilde p(w_1,...,w_n|\alpha, \beta)$ from $\BE[\theta(z')|w_1,...,w_n]$ as a function of $\epsilon$,
using the computation of the previous proof.

\section{Conclusions and outlook}
\label{sec:Outlook}
We showed that by using generating power series techniques \eqref{eq:def_p} can be computed efficiently (in linear time in the number of causes) for a small (fixed) number of observations, or, more generally, for a small tree width of the interaction graph. The given algorithm is optimal for an online algorithm, and even without the ``online'' condition there is no exact algorithm that is polynomial in both number of observations and number of causes unless $P=NP$.\CR
So for larger (but still interesting, not too large) numbers of observations (and a larger tree width of the interaction graph) we have to resort to approximation algorithms. Of course there already are approximation algorithms in use (e.g. Variational Bayes or Gibbs Sampling), but it would be interesting to compare them to approximations that arise from the formulas established here by clustering ``similar'' observations and approximate them by using the same number of artificial, averaged ``cluster observations'', such that we have a smaller number of different ``cluster observations''. These artificial observations are then also more dissimilar and we can apply the tree width computation above as an approximation by neglecting products that involve ``very dissimilar'' observations.\CR
\clearpage

\clearpage
\appendix
\appendixpage
\section{Mathematical notations}
{\bf Partitions:}\CR
A partition $\pi$ of a set $I$ is a set of disjoint subsets $\pi=\{I_1,...,I_k\}$ of $I$ such
that $I=I_1 \cup...\cup I_k$. We denote by $Part(I)$ the set of all partitions of $I$.
\smallspace
{\bf Commutative ring:}\CR
A ring $(R,+,\cdot)$ is a set $R$ on which addition and multiplication are defined such that
some axioms are satisfied (see any text book on Algebra, but for us the examples below are enough). 
All rings we use will be commutative ($f\cdot g = g\cdot f$).\CR
Examples are the integers $\BZ$, or $\BR$, or for any commutative ring $R$ (in particular $R=\BR$) 
the polynomials in one variable $R[X]$,
or polynomials in several variables $R[X_1,...,X_n]$. Below we will also introduce the commutative ring of 
formal power series in one or more variables $R[[X_1,...,X_n]]$.
\smallspace
{\bf Ideal, modulo:}\CR
An ideal $\calI$ in a commutative ring $R$ is a subset $\calI\subset R$ which is closed under addition and under
multiplication with elements of $R$. Examples are for each natural number $m$ the ideal $(m) := m\BZ \subset \BZ$ 
or for each natural number $r$ the set of all polynomials $(X^r) := X^r \cdot \BR[X] \subset \BR[X]$.\CR
More generally, for any $r_1,...,r_n\in R$ we denote by $(r_1,...,r_n)$ the set of all ring elements that can 
be written as ``linear combination'' of the $r_i$, i.e.
\[
   (r_1,...,r_n) := \{ \lambda_1\cdot r_1 + ... + \lambda_n\cdot r_n \ |\ \lambda_1,...,\lambda_n\in R\}.
\]
This is an ideal in $R$, called the ideal generated by $r_1,...,r_n$.
\CR
Given an ideal $\calI\subset R$, we can compute ``modulo $\calI$'': We write
\[ 
   f \equiv g \mod \calI  \qquad:\Leftrightarrow f-g\in \calI
\]
This notion satisfies the usual rules that are also satisfied for computations modulo $m$ in $\BZ$:
The relation $\equiv$ is an equivalence relation and if $f\equiv f'$ and $g\equiv g'$, 
we also have $f+g \equiv f'+g'$ and $f\cdot g \equiv f'\cdot g'$.
As a consequence, the set $R/\calI$ of equivalence classes is again a commutative ring 
(generalizing the rings $\BZ/m$ of integers modulo $m$).
\smallspace
{\bf Algebra over $\BR$:}\CR
An algebra over $\BR$ is a $\BR$-vector space on which a multiplication (of ``vectors'') is defined, such that
the axioms of a ring are satisfied. Examples are polynomials $\BR[X_1,...,X_n]$, power series $\BR[[X_1,...,X_n]]$,
and $\BR[X_1,...,X_n]/(X_1^{k_1},...,X_n^{k_n})$ for natural numbers $n,k_1,...,k_n$. 
The dimension of the vector space is also called
dimension of the algebra --- so the first two examples are infinite--dimensional algebras, but 
$\BR[X_1,...,X_n]/(X_1^{k_1},...,X_n^{k_n})$ is a $k_1\cdot ... \cdot k_n$--dimensional algebra. Elements in this 
algebra are given by (the equivalence classes of) polynomials 
\[
   \sum_{i_1 =0}^{k_1-1} \cdots \sum_{i_n =0}^{k_n-1} c_{i_1,...,i_n}\cdot X_1^{i_1} \cdots X_n^{i_n}
\] 
which in turn are given by the $d=k_1\cdot ... \cdot k_n$ numbers $c_{i_1,...,i_n}$.
Two such elements of this $d$-dimensional algebra can be added with $d$ addition operations, and multiplied 
with at most $d^2$ multiplications and additions.
\smallspace
{\bf Formal power series:}\CR
A formal power series over a commutative ring $R$ (we will use $R=\BR$ most of the time) is 
a formal infinite series
\[
   a_0 + a_1\cdot X + a_2\cdot X^2 + ...
\]
with $a_i\in R$. The usual symbolic addition / multiplication make the formal power series into a ring
$R[[X]]$. Similarly, we can define the ring of formal power series in several variables $R[[X_1,...,X_n]]$.
Apart from addition and multiplication we can also take derivatives $\partial/\partial X_i$ and substitute a power series
without constant term into another, these are computations
that can be described purely symbolically, hence they are also defined for formal power series.
\CR
If we can express a function $\BR^n\arrow \BR$ by a power series in $\BR[[X_1,...,X_n]]$
that is convergent in a neighborhood of $(0,...,0)$, the coefficients of this power series are determined by the function,
and we will use this function also to denote the corresponding power series.
For example, since the series
\begin{eqnarray*}
    1/(1+X) &=& 1 - X + X^2 -+ ...\\
    \exp(X) &=& 1 + \frac{X}{1} + \frac{X^2}{2!} + \frac{X^3}{3!} + ... \\
    \log(1+X) &=& X - X^2 / 2 + X^3/3 -+ ...
\end{eqnarray*}
are convergent for all $X\in\BR$ with $|X|<1$, these functions determine the coefficients and we will also consider
them as elements of $\BR[[X]]$. Since we 
have e.g. $\exp(\log(1+X)) = 1+X$ and $\exp(X+Y) = \exp(X)\cdot \exp(Y)$ as convergergent series,
such identities also hold for the formal power series in $\BR[[X,Y]]$. As a consequence of the first equation we
can see that each power series in $\BR[[X_1,...,X_n]]$ with constant term $\neq 0$ has a multiplicative inverse, i.e. we
can also divide by such power series.
\smallspace
{\bf Pochhammer symbol:}\CR
For $\alpha \in \BR$ and $i$ a natural number we set
\[
   (\alpha)_i := \alpha \cdot (\alpha+1) \cdot ... \cdot (\alpha+i-1)
\]
For $i=0$ we set $(\alpha)_0 := 1$.\CR
With this notation we have the Taylor series
\begin{eqnarray}
 \lefteqn{(1-X)^{-\alpha}} \nonumber\\
  &=& 1 + \alpha x + \frac{\alpha(\alpha+1)}{2!} X^2 
         + \frac{\alpha(\alpha+1)(\alpha+2)}{3!} X^3+... \nonumber\\
  &=& \sum_{i=0}^\infty (\alpha)_i \frac{X^i}{i!}
   \label{eq:Pochhammer}
 \end{eqnarray}
This series converges for $|X|<1$, so again this function determines the coefficients and we 
write $(1-X)^{-\alpha}$ also for the corresponding formal power series. Since
\[
   (1-X)^{-\alpha} = \exp\big(-\alpha\cdot \log(1-X)\big)
\]
as functions, this is also true as equation between formal power series.
\section{Toy example}
\subsection{Maximum likelihood vs. marginal inference}
Our toy example has 3 ``causes'' and 2 ``observations'', with the following probabilities $\beta$:
\smallspace
\begin{tabular}{r||l|l|l}
  &$z_1$ & $z_2$  & $z_3$ \\
  \hline
  $\beta(w_1|z)$& 0.09 & 0.05 & 0.02 \\
  $\beta(w_2|z)$& 0.02 & 0.05 & 0.08
\end{tabular}
\smallspace
The likelihood function is 
\begin{eqnarray*}
   p(w_1,w_2|\theta, \beta) &=& \prod_{w}\sum_{z} \beta(w|z)\theta(z) \\
   &=& (0.09\theta_1 + 0.05 \theta_2 + 0.02\theta_3) 
   \cdot (0.02\theta_1 + 0.05 \theta_2 + 0.08\theta_3),
\end{eqnarray*}
it is defined in the plane in $\BR^3$ given by $\theta_1 + \theta_2 + \theta_3=1$ where
$0\leq \theta_1, \theta_2, \theta_3 \leq 1$. It looks like this:
\CR
\hskip 5mm\includegraphics[width=0.4\textwidth]{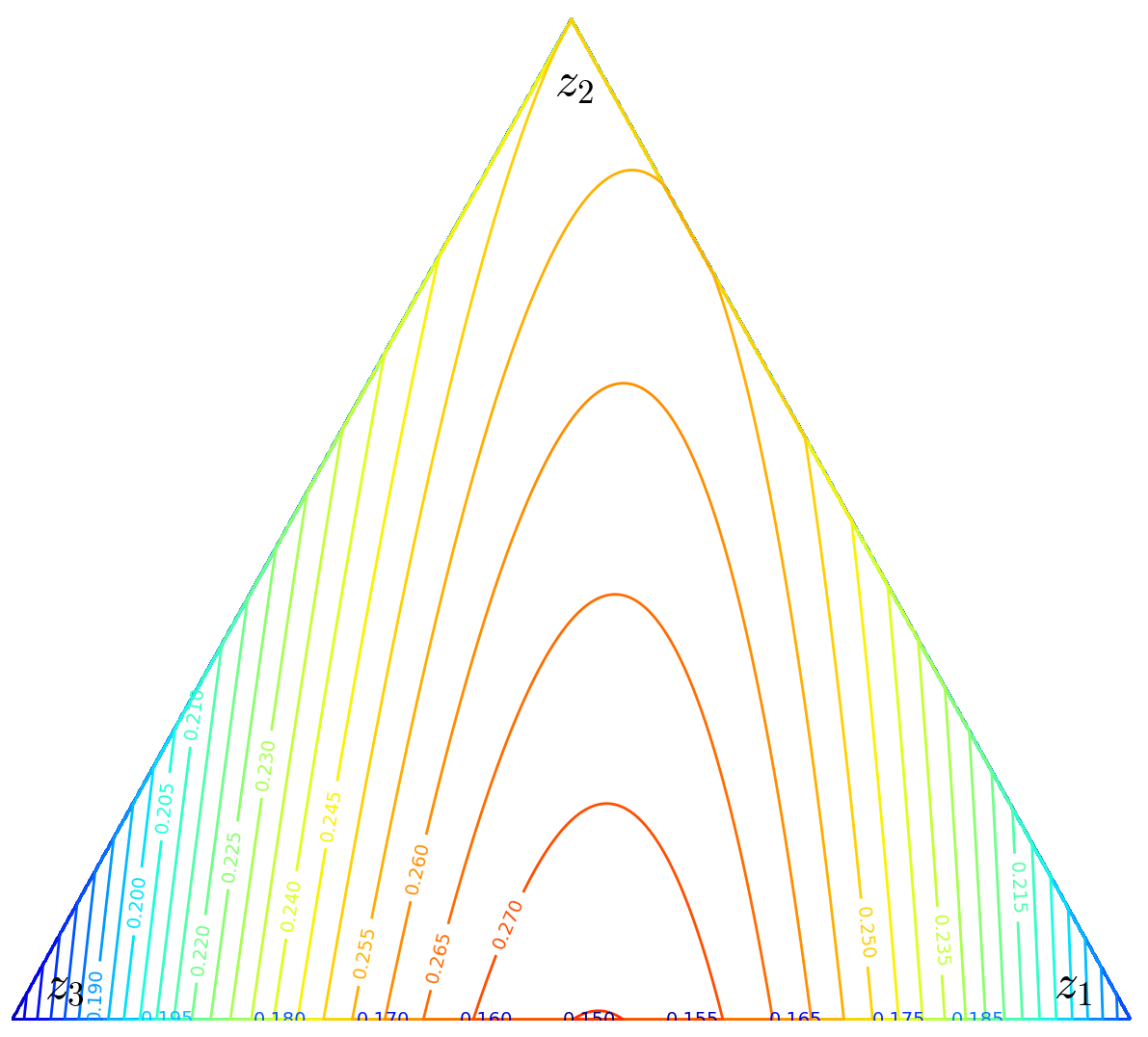}
\CR
The maximum likelihood solution has $\theta_1$ and $\theta_3$ around 1/2, and $\theta_2=0$, but there
are almost as likely solutions for all values of $\theta_2$, up to $\theta_2=1$, so it would be misleading
if we only looked at the maximum likelihood solution and concluded that ``cause'' $z_2$ did not contribute
to our observations.
\smallspace
To contrast this with the marginal inference formula, we have to specify a prior $Dir(\alpha)$
on $\theta$. We will use the uniform prior, it is given by $\alpha=(1,1,1)$.
For the uniform prior, the posterior distribution is proportional to 
likelihood of $\theta$:
\[
   p(\theta|w,\alpha,\beta) = p(w|\theta,\beta) \cdot \frac{p(\theta|\alpha)}{p(w|\alpha,\beta)}
\]
So the maximum a posteriori solution is the maximum likelihood solution, and the marginal inference
is the average $\theta$ under the posterior distribution, this actually gives $z_2$ the largest weight:
\smallspace
\begin{tabular}{r||l|l|l}
  &$z_1$ & $z_2$  & $z_3$ \\
  \hline
  $\alpha(z)$ & 1 & 1 & 1 \\
  \hline 
  Maximum likelihood & 0.524 & 0 & 0.476 \\
  Exact Bayes &  0.335 & 0.337 & 0.327   \\
\end{tabular}

\subsection{Variational Bayes}
In the following we keep $\alpha, \beta$ fixed and drop it from the notation.\CR
In the variational approximation of \cite{BleiNgJordan}, the distribution $p(\theta,z|w)$ is 
approximated by a distribution of the form 
\begin{eqnarray*}
   q(\theta,z|\gamma, \phi) &=& q(\theta|\gamma) \cdot q(z|\phi) \\
     &=& q(\theta|\gamma)\cdot q(z_1|\phi_1)\cdot...\cdot q(z_n|\phi_n),
\end{eqnarray*}
where $\gamma$ is a parameter for the Dirichlet distribution $\theta\sim Dir(\gamma)$ 
given by the $m$ numbers $\gamma(z)$ for $z\in Z$, and the $\phi_i$ are multinomial parameters, 
given by the $n\times m$ probabilities $\phi_i(z)$ for $z\in Z$.
\CR
Given $\alpha,\beta$ and the observations $w$, the parameters $\gamma,\phi$ are chosen to 
minimize the Kullback--Leibler divergence
\[
    D\Big(p(\theta,z|w) \,\Big\|\, q(\theta,z|\gamma,\phi)\Big)
\]
The iterative algorithm of \cite{BleiNgJordan}, p.1005 computing these parameters is:
\CR
Start with $\gamma(z) := \alpha(z) + n/m$ and iterate
\begin{eqnarray*}
  \phi_i(z) &\propto& \beta(w_i|z) \cdot e^{\Psi(\gamma(z)) - \Psi\left(\sum_{z'} \gamma(z')\right)}\\
  \gamma(z) &=& \alpha(z) + \sum_{i=1}^n \phi_i(z)
\end{eqnarray*}
until convergence. (Here $\Psi$ is the digamma function, i.e. the logarithmic derivative of $\Gamma(x)$.)
\CR
We are interested in the distribution $Dir(\gamma)$ which approximates the posterior 
distribution $p(\theta|w)$, its mean $\gamma / (\sum_z \gamma(z))$ gives the approximation to 
our marginal inference. In our example we get
\smallspace
\begin{tabular}{r||l|l|l}
  &$z_1$ & $z_2$  & $z_3$ \\
  \hline
  $\alpha(z)$ & 1 & 1 & 1 \\
  \hline 
  Maximum likelihood & 0.524 & 0 & 0.476 \\
  Variational Bayes & 0.344 & 0.324 & 0.331 \\
  Exact Bayes &  0.335 & 0.337 & 0.327   \\
\end{tabular}
\smallspace
Note, however, that the variational Bayes solution is derived from the posterior distribution 
$p(\theta, z | w, \alpha, \beta)$ of both the general mixture $\theta$ of the causes, and the particular 
causes $z$ that gave rise to the observations $w$. The simplifying assumption that the 
distribution of the $z$ is independent of $\theta$ is a good approximation to the reality if the
distribution of the $\theta$ is concentrated around one point, i.e. we are quite sure about the underlying
distribution of causes --- this is usually the case if we have many observations or a strong prior. 
But when this is not the case, there will be no good fit in the variational family, and then the 
$Dir(\gamma)$ part can also be quite different from
the best fit to the posterior distribution of $\theta$.
We can see an example of that when we reduce the prior to $\alpha=(1/3, 1/3, 1/3)$:
\smallspace
\begin{tabular}{r||l|l|l}
  &$z_1$ & $z_2$  & $z_3$ \\
  \hline
  $\alpha(z)$ & 1/3 & 1/3 & 1/3 \\
  \hline 
  Maximum likelihood & 0.524 & 0 & 0.476 \\
  Variational Bayes & 0.446 & 0.151 & 0.403 \\
  Exact Bayes &  0.331 & 0.355 & 0.314   \\
\end{tabular}
\subsection{Concentration bias}
To see more concretely how this difference arises, we can apply the chain rule for the
Kullback--Leibler divergence
\begin{eqnarray}
  \lefteqn{ D\Big( q(\theta,z) \,\big\|\, p(\theta,z) \Big)}  \label{eq:chainRule} \\
  &=& D\Big( q(\theta) \,\big\|\, p(\theta) \Big) + \BE_{\theta\sim q} 
     D\Big( q(z|\theta) \,\big\|\, p(z|\theta) \Big) \nonumber
\end{eqnarray}
to the true posterior distribution $p(\theta, z|w)$ and its approximation $q(\theta,z|\gamma, \phi)$.
The first summand of \eqref{eq:chainRule} is the KL divergence between the approximation 
$q(\theta)$ and our true posterior $p(\theta|w)$ that we ideally would be minimizing (but this is 
difficult because $p(\theta|w)$ involves the likelihood function). The second summand gives the 
average KL divergence between $q(z|\theta)$, which is given by $\phi$, independent of $\theta$,
and 
\[
   p(z|\theta,w) = \prod_{i=1}^n \beta(w_i|z_i)\cdot \frac{\theta(z_i)}{\sum_z \beta(w_i|z) \cdot \theta(z)} 
\]
which varies with the $\theta$.
\CR
So this second term favors those $q(\theta|\gamma)$ that are concentrated in a small area, which can result
in $q(\theta|\gamma)$ that are not the best approximations to $p(\theta|w)$.
\CR
In the first example with the (uniform) prior $\alpha=(1,1,1)$ this does not affect the average much:
\smallspace
\includegraphics[width=0.3\textwidth]{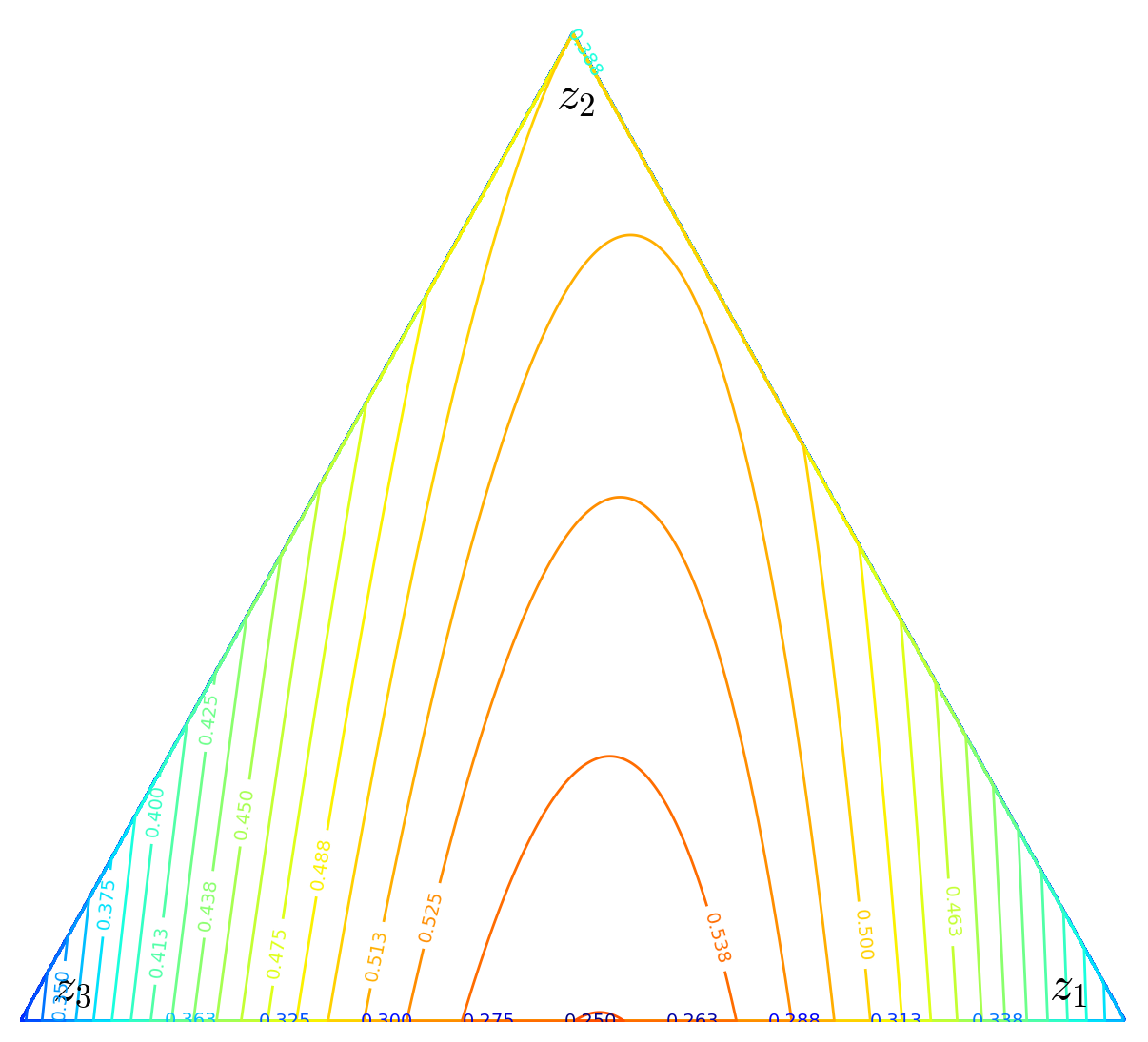}
\includegraphics[width=0.3\textwidth]{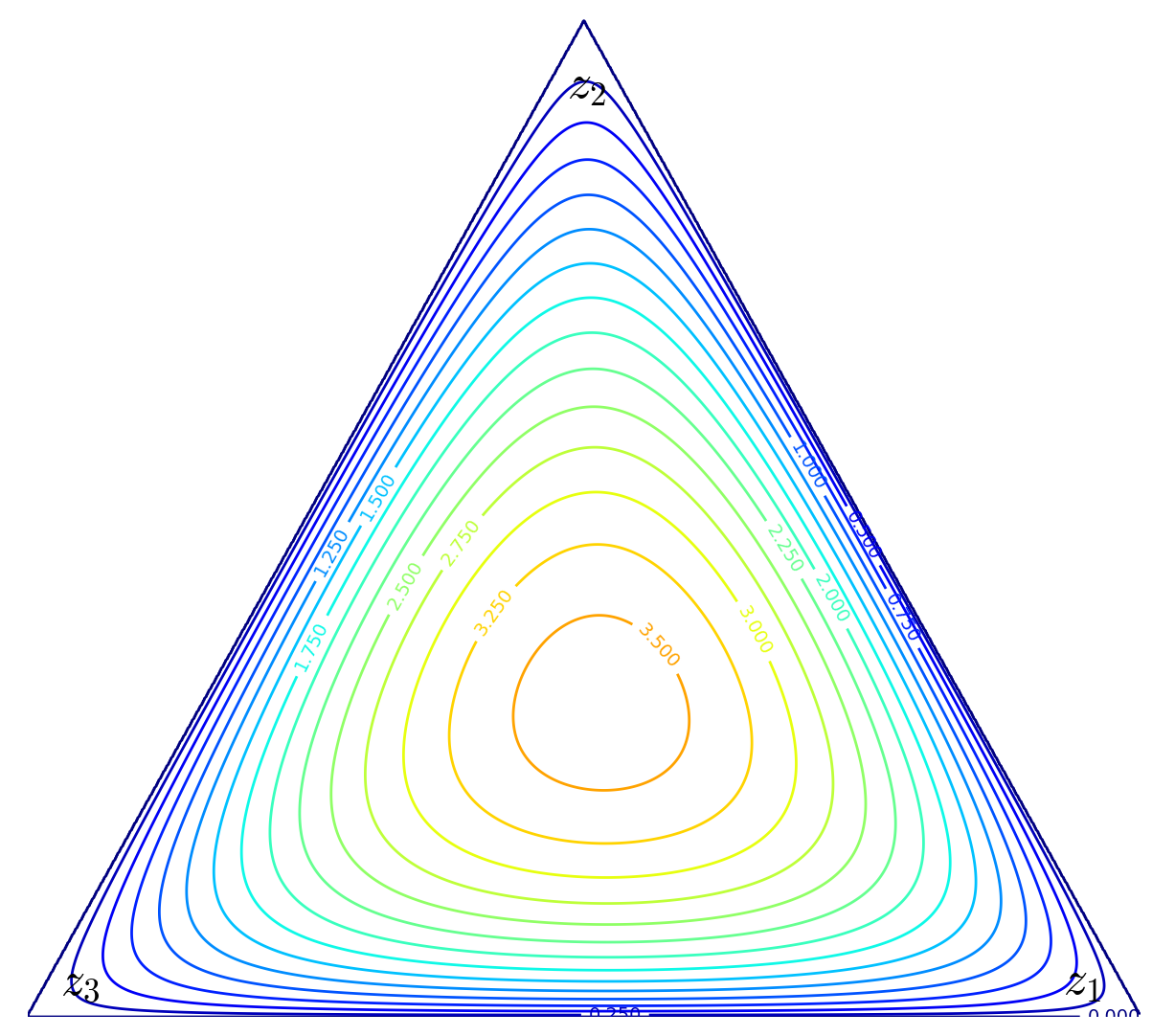}\CR
\hskip 14mm $p(\theta|w)$ \hskip 26mm $q(\theta|\gamma)$
\smallspace
But for the prior $\alpha=(1/3, 1/3, 1/3)$ the posterior $p(\theta|w)$ is largest around the
edges of the triangle, and the approximation $q(\theta|\gamma)$ focuses only on the lower side,
which has a higher likelihood than the other two sides:
\smallspace
\includegraphics[width=0.3\textwidth]{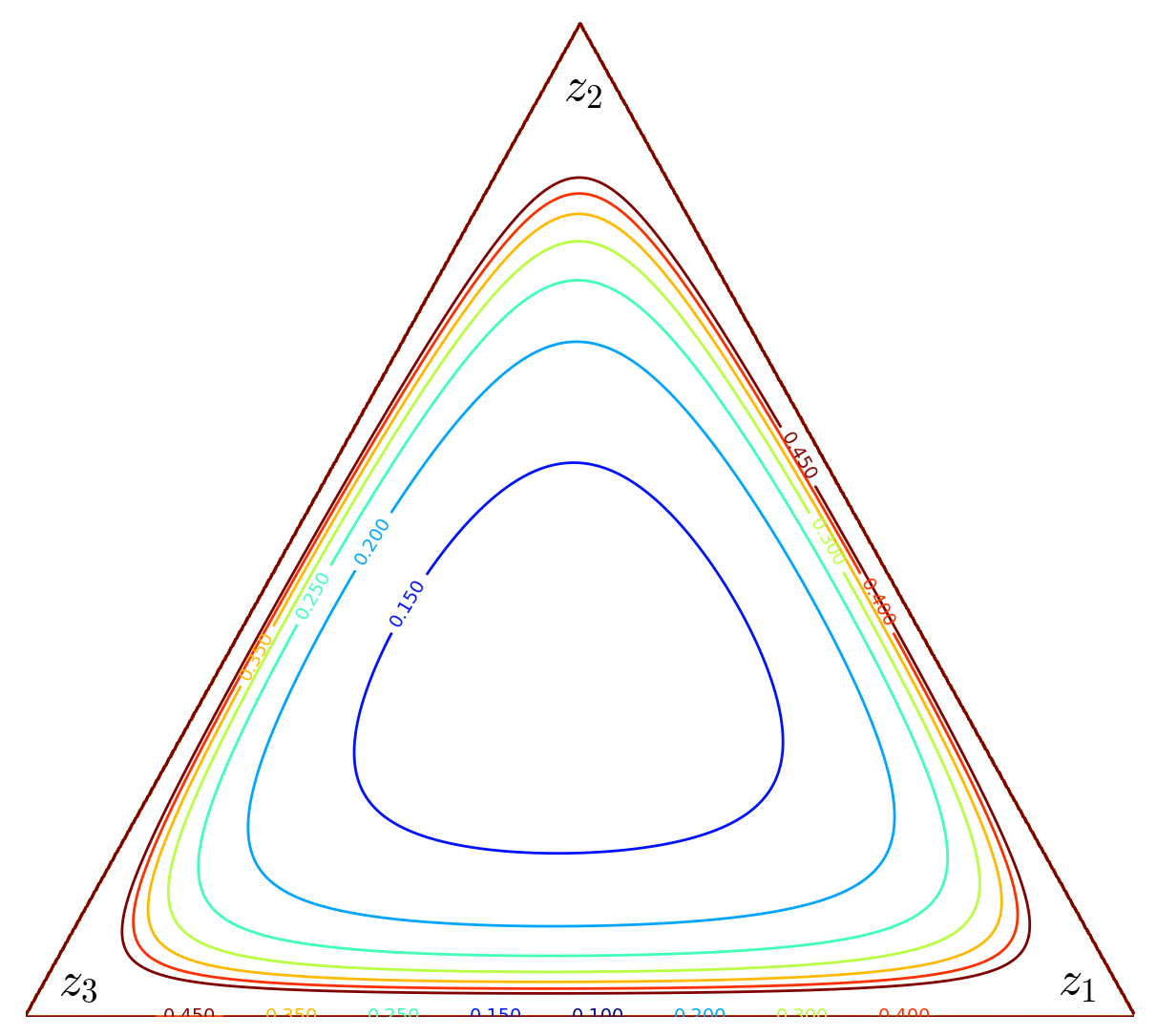}
\includegraphics[width=0.3\textwidth]{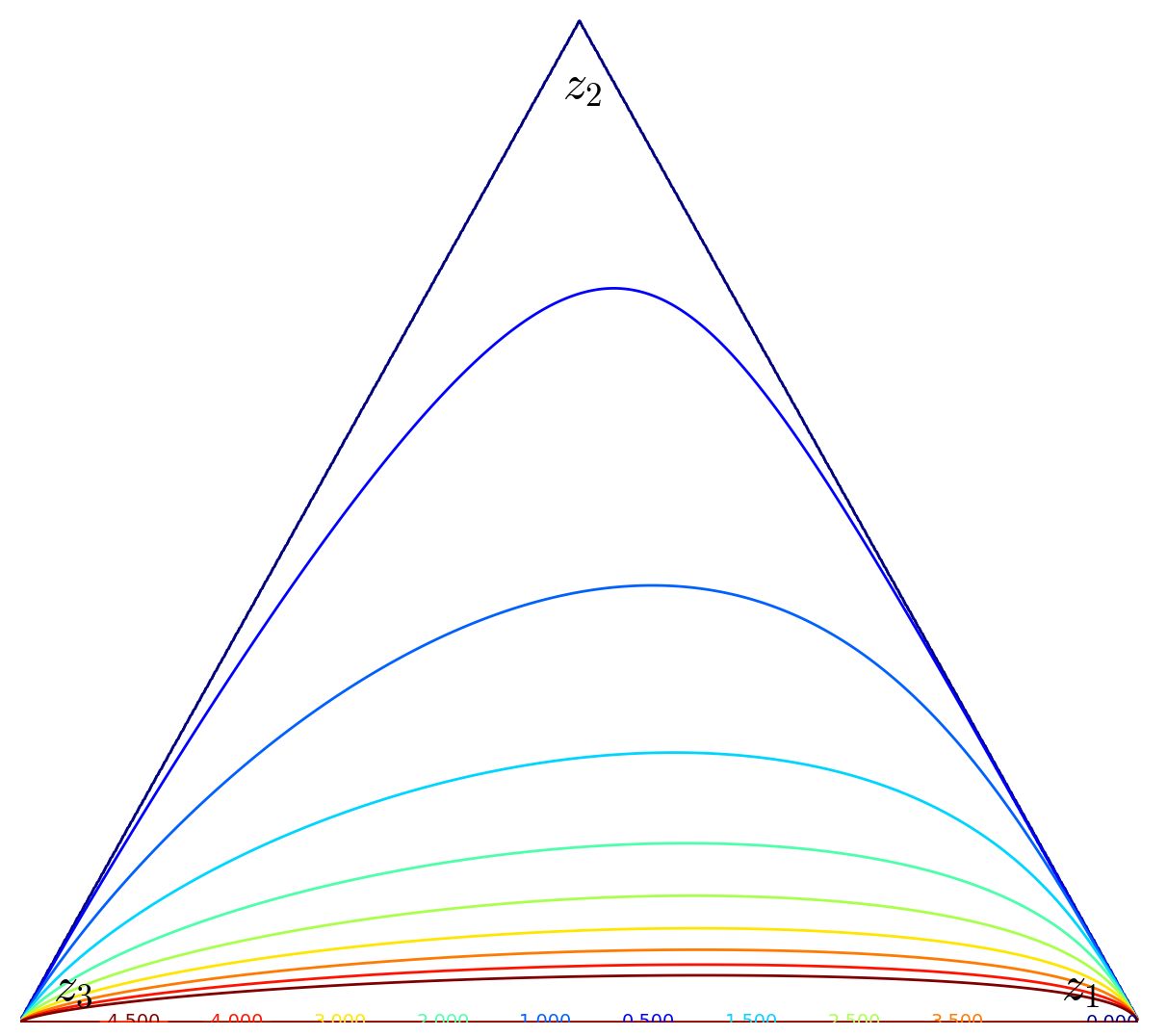}\CR
\hskip 14mm $p(\theta|w)$ \hskip 26mm $q(\theta|\gamma)$

\subsection{Subdivisions of the causes}
There is another issue with using Variational Bayes for a large number of 
causes like locations: We may look at locations in different granularity, and
expect that e.g. the probability of a cause $z$ lying in a country is the sum of 
the probabilities of $z$ lying in the country's provinces. This is true for
the exact solution, but not for the Variational approximation:
\smallspace
The marginals of Dirichlet distributions are again Dirichlet distributions (see e.g. 
\cite{GhoshRamamoorthi}, p.90):
When we have a prior distribution of mixtures 
\begin{eqnarray*}
  \theta' \sim Dir(\alpha') & & \hbox{with}\\
     \theta' &:=& (\theta_{1a}, \theta_{1b}, \theta_2, \theta_3,...,\theta_m) \\
     \alpha' &:=& (\alpha_{1a}, \alpha_{1b}, \alpha_2, \alpha_3, ...,\alpha_m)
\end{eqnarray*}
the marginal distribution defined by merging the first two causes $z_{1a}, z_{1b}$ to one 
cause $z_1$ gives
\begin{eqnarray*}
    \theta \sim Dir(\alpha) & & \hbox{with} \\
     \theta &:=& (\theta_{1a}+\theta_{1b}, \theta_2,...,\theta_m) \\
     \alpha &:=& (\alpha_{1a}+\alpha_{1b}, \alpha_2, \alpha_3, ...,\alpha_m)
\end{eqnarray*}

If also $\beta(w|z_{1a}) = \beta(w|z_{1b}) =: \beta(w|z_1)$ for all $w$, we have essentially the same generative process
if we either choose 
\begin{itemize} 
  \item[a)] $\theta'\sim Dir(\alpha')$, $z'\sim \theta'$, $w\sim \beta(.|z')$, or
  \item[b)] $\theta\sim Dir(\alpha)$, $z\sim \theta$, $w\sim \beta(.|z)$
\end{itemize}
the only difference is that a $z_1$\ in b) can correspond to either a $z_{1a}$ or $z_{1b}$\ in a). 
\CR
So if in our example we subdivide the first ``location area'' $z_1$ into two smaller areas $z_{1a}$ and $z_{1b}$, 
we would expect that this does not change the probabilities $\theta(z_2), \theta(z_3)$ 
of the other locations, and 
also $\theta(z_1)$ = $\theta(z_{1a}) + \theta(z_{1b})$.
\smallspace
However, we can subdivide the first cause in our toy example into two equally likely subcauses,
and see that for the Variational Bayes approximation this invariance is clearly not satisfied:
\smallspace
\begin{tabular}{r||l|l|l|l}
  &$z_{1a}$ &  $z_{1b}$ & $z_2$  & $z_3$ \\
  \hline
  $\beta(w_1|z)$& 0.09 & 0.09 & 0.05 & 0.02 \\
  $\beta(w_2|z)$& 0.02 & 0.02 & 0.05 & 0.08 \\
  $\alpha(z)$ & 1/6 & 1/6 & 1/3 & 1/3 \\
  \hline
  Variational Bayes & 0.056 & 0.056 & 0.741 & 0.147 \\
  Exact Bayes & 0.165 &  0.165 & 0.355 & 0.314  \\
\end{tabular}

\subsection{Gibbs sampling}
Gibbs sampling does not have these issues, but is 
slower and non-deterministic:\CR
To sample from the posterior $p(\theta|w,\alpha,\beta)$, we alternately:
\begin{itemize}
  \item Sample $z_i$ such that $p(z_i) \propto \theta \cdot \beta(w_i|.)$
  \item Sample $\theta \sim Dir(\alpha + [z])$, where $[z]$ has for each cause
     the number of times this causes appears in the sampled $z_1,...,z_n$.
\end{itemize}
and output the average $\theta$.

In ten runs with one million iterations I got in the mean (and $\pm...$ gives
the observed standard deviation of these values)
\smallspace
\begin{tabular}{r||l|l}
     &  our formula & Gibbs sampling \\
  \hline
 $z_1$ & 0.3309 & $0.3313 \pm 0.0005$ \\
 $z_2$ & 0.3549 & $0.3549 \pm 0.0005$ \\
 $z_3$ & 0.3141 & $0.3138 \pm 0.0005$ 
\end{tabular}
\smallspace
After subdivision of the first area/cause we get:\CR
\begin{tabular}{r||l|l}
     &  our formula & Gibbs sampling \\
  \hline
 $z_1$ & 0.1655 & $0.1654 \pm 0.0005$ \\
 $z_2$ & 0.1655 & $0.1658 \pm 0.0004$ \\
 $z_3$ & 0.3549 & $0.3551 \pm 0.0004$ \\
 $z_3$ & 0.3141 & $0.3137 \pm 0.0006$ \\
\end{tabular}
\CR

\section{Proof of Lemma \ref{DirichletGamma}}
\begin{namedthm}{Lemma \ref{DirichletGamma}}
Let $f:\BR_{>0}^m\arrow \BR$ be a function that is homogeneous of degree $h$, i.e. 
\[
   f(t\cdot \theta) = t^h\cdot f(\theta)
   \qquad \hbox{for} \quad
   t > 0
\]
then 
\[
   \BE_{\theta\sim Dir(\alpha)}[f(\theta)] = 
   \frac{\Gamma(|\alpha|)}{\Gamma(|\alpha|+h)} \BE_{\theta\sim \Gamma(\alpha,1)}[f(\theta)]
\]
\end{namedthm}
\noindent
{\bf Proof:}\CR
\begin{eqnarray*}
\BE_{\theta\sim Dir(\alpha)}[f(\theta)]
  &=& \int_{\theta\in\Delta} \frac{\theta^{\alpha-1}}{B(\alpha)} f(\theta) d\theta \\
  &=& \frac{\int_0^\infty \int_{\theta\in\Delta} (t\theta)^{\alpha-1}e^{-t} f(t\theta)\cdot t^{m-1} d\theta dt}
       {B(\alpha) \cdot \int_0^\infty t^{|\alpha| -m + h + m-1} e^{-t} dt} \\
  &=& \frac{1}{B(\alpha)\cdot \Gamma(|\alpha| + h)}
       \int_{\theta\in\BR_{>0}^m} \theta^{\alpha-1} e^{-|\theta|} f(\theta) d\theta \\
  &=& \frac{\Gamma(\alpha)}{B(\alpha)\cdot \Gamma(|\alpha| + h)}
       \BE_{\theta\sim\Gamma(\alpha,1)}[f(\theta)] \\
  &=& \frac{\Gamma(|\alpha|)}{\Gamma(|\alpha| + h)}
       \BE_{\theta\sim\Gamma(\alpha,1)}[f(\theta)]
\end{eqnarray*}
\QEDA

\section{Proof of Lemma \ref{GenerateS}}
\begin{namedthm}{Lemma \ref{GenerateS}}
\begin{eqnarray}
   S(X) 
   &:=& \sum_{I\in \calM(W)} \ \tilde{p}(I|\alpha,\beta) \cdot \frac{X^I}{I!} \nonumber \\
   &=&  \prod_{j=1}^m 
    \Big(1-\sum_{i=1}^r \beta(\tilde{w}_i|z_j)\cdot X_i\Big)^{-\alpha_j}
    \label{eq:product_power_alphaA}
\end{eqnarray}
\end{namedthm}
\noindent
{\bf Proof:}\CR
(We assume $\alpha,\beta$\ fixed and don't include it in the notation introduced here)
\CR
For a sequence of observations $(\tilde w_1,...,\tilde w_r)$ we introduce the notation
\[
   q_\theta\big(Z,(\tilde w_1,...,\tilde w_r)\big) := \sum_{z\in Z^n} \prod_{i=1}^r \beta(\tilde w_i|z_i) \cdot \theta(z_i)
\]
and will also abbreviate this as
\[ 
   q_\theta\big(Z,I\big) := q_\theta\big(Z,(\tilde w_1,...,\tilde w_r)\big)
\]
for the multiset $I$ given by this sequence $(\tilde w_1,...,\tilde w_r)$.\CR
We will use the notation $I+J$ for the ``union'' of multisets (corresponding to adding up the multiplicities), and 
$\binom{J+K}{K}= (J+K)!/(J!\cdot K!)$. 
Then if we have a partition
$Z = Z' \cup Z''$ of $Z$ into two disjoint parts, we have
\begin{equation}
    q_\theta(Z'\cup Z'', I)
    = \sum_{J+K=I} \binom{I}{J} \cdot q_\theta(Z',J) \cdot q_\theta(Z'',K)
    \label{eq:convolution}
\end{equation}
since once $J$ is chosen, i.e. we have decided that we want $j_k$ of the $i_k$ observations $\tilde{w}_k$ to come from $Z'$,
there are $\binom{i_k}{j_k}$ ways to choose which of the $i_k$ observations to assign to a cause in $Z'$.\CR
Now introduce the formal power series
\[
   Q_{\theta,Z}(X) := \sum_{I\in\calM(W)} q_\theta(Z,I) \cdot \frac{X^I}{I!}
\]
then we can reformulate (\ref{eq:convolution}) as 
\begin{equation}
   Q_{\theta,Z'\cup Z''}(X) = Q_{\theta,Z'}(X) \cdot Q_{\theta,Z''}(X) \label{eq:multS}
\end{equation}
Then for the full set $\{z_1,...,z_m\}$ of possible causes we have the factorization
\[
   S(X) = \BE_{\theta\sim\Gamma(\alpha,1)} [Q_{\theta,\{z_1,...,z_m\}}]
        = \prod_{j=1}^m \BE_{\theta\sim\Gamma(\alpha_j,1)} [Q_{\theta,\{z_j\}}]
\]
and this allows us to decompose
this expectation into $m$ simpler factors, each involving only one cause $z$. 
If in $I$ the words $w_i$ appear with multiplicity $k_i$, we have
\[
  q_\theta(\{z\},I)
     = \big(\beta(w_1|z)\cdot\theta(z)\big)^{k_1} 
       \cdot ... \cdot 
       \big(\beta(w_n|z)\cdot\theta(z)\big)^{k_r}     
\]
which gives
\begin{eqnarray*}
  Q_{\theta,\{z\}}(X)
   &=&\exp\big(\beta(\tilde{w}_1|z)\theta(z)\cdot X_1\big)
                     \cdot...\cdot
                     \exp\big(\beta(\tilde{w}_r|z)\theta(z)\cdot X_r\big) \\
   &=&\exp\Big(\sum_{i=1}^r \beta(\tilde{w}_i|z)\theta(z)\cdot X_i \Big)
\end{eqnarray*}
With
\[
   Y := \sum_{i=1}^r \beta(\tilde{w}_i|z)\cdot X_i
\]
we get with the moment generating function of the Gamma distribution
\begin{eqnarray*}
   \BE_{\theta\sim\Gamma(\alpha,1)} \left[Q_{\theta,\{z\}}(X)\right] 
    &=&  \BE_{\theta_z\sim\Gamma(\alpha_z,1)} \left[e^{\theta_z\cdot Y}  \right] \\
    &=& (1-  Y)^{-\alpha_z} \\
    &=& (1- \sum_{i=1}^r \beta(\tilde{w}_i|z)\cdot X_{w_i})^{-\alpha_z}
\end{eqnarray*}
which in turn gives the formula \eqref{eq:product_power_alphaA} because of the factorization.
\QEDA
\smallspace
One can check that all used transformations are valid for formal power series, so we do not have 
to prove convergence. However, in this case all power series are even convergent in a 
neighborhood of (0,0,...,0), so the computations are also valid as analytic functions in this domain.

\section{Proof of Lemma \ref{pExplicit}}
\begin{namedthm}{Lemma \ref{pExplicit}}
All coefficients $\tilde p(I|\alpha,\beta)$ of $X^I$ for $I$ a set 
(i.e. all multiplicities $\leq 1$) in the power series $S(X)$ 
agree with the coefficients of the polynomial
\begin{equation}
  \prod_{\emptyset \neq J\subseteq W}
        \Big(1 + \langle \beta_J \rangle\cdot \Gamma(|J|)\cdot X^J\Big)
  \label{eq:AllDiffWordsProd2}
\end{equation}
Equivalently, we can write $\tilde p(I|\alpha,\beta)$ for a set $I$ as a sum over all partitions $\pi$ of $I$ into subsets:
\begin{equation}
   \tilde{p}(I|\alpha, \beta) = \sum_{\pi\in Part(I)} \prod_{J\in\pi} \ 
   \langle \beta_J \rangle \cdot \Gamma(|J|)
   \label{eq:SimpleFormulaA}
\end{equation}
\end{namedthm}
\noindent
{\bf Proof:}\CR
We get for  the logarithm of \eqref{eq:product_power_alpha} at one cause $z$
\begin{eqnarray*}
  -\alpha \cdot \log(1-\sum_{i=1}^r \beta_i X_i)
  &=& \alpha \cdot\big(\sum_{i=1}^r \beta_i\ X_i\big)
    + \alpha \cdot\big(\sum_{i=1}^r \beta_i\ X_i\big)^2 / 2
    + ... \\
  &=& \sum_{\emptyset\neq J\in\calM(W)} \alpha \cdot \beta_J \cdot \frac{|J|!}{J!} 
      \cdot \frac{X^J}{|J|} \\
  &=& \sum_{\emptyset\neq J\in\calM(W)} \alpha \cdot \beta_J \cdot \Gamma(|J|)
      \cdot \frac{X^J}{J!}
\end{eqnarray*}
So if we use the abbreviation
$
   \langle \beta_J \rangle := \sum_z \alpha(z) \beta_J(z)
$
we get for the generating function
\begin{eqnarray}
 S(X)
   &=& \exp\Big(\sum_{z\in Z} \sum_{J\in \calM(W)} \alpha(z) \beta_J(z) \Gamma(|J|)
          \cdot \frac{X^J}{J!}\Big)\nonumber  \\
   &=& \exp\Big( \sum_{J\in \calM(W)} \langle \beta_J \rangle \Gamma(|J|)
          \cdot \frac{X^J}{J!}\Big)\nonumber  \\
   &=& \prod_{J\in \calM(W)} \exp\Big(\langle \beta_J \rangle\cdot \Gamma(|J|)
          \cdot \frac{X^J}{J!}\Big)   \label{eq:GeneralExpProd} 
\end{eqnarray}
If we discard all $X^J$ which contain a $X_w^2$ the product becomes just
\[
 S(X) \equiv\prod_{\emptyset \neq J\subseteq W}
        \Big(1 + \langle \beta_J \rangle\cdot \Gamma(|J|)\cdot X^J\Big)
        \mod \big( X_1^2,...,X_r^2\big)
\]
Multiplying out we get the explicit formula for $\tilde p(I|\alpha, \beta)$.
\QEDA

\section{Proof of Theorem 1}
{\bf Theorem 1:}
The $p(w_1,...,w_n|\alpha,\beta)$ can be computed in time $O(3^n + m\cdot 2^n)$ and space $O(2^n)$.
\CR
{\bf Proof:}
We compute the generating power series \eqref{eq:AllDiffWordsProd} in the algebra
\[
    A := \BR[X_1,...,X_n] / (X_1^2,...,X_n^2)
\]
of polynomials in $n$ variables modulo the ideal generated by the squares of the variables. This is an algebra of degree $2^n$ over $\BR$, a basis is given by the $X^J$ where the $J$ are subsets of $W$ (equivalently, multisets with multiplicities $\leq$ 1).
 To compute $\tilde{p}(W)$, we have first to compute the $2^n$ numbers 
$\langle \beta_J \rangle$, which takes $O(m\cdot 2^n)$ time, then compute the product \eqref{eq:AllDiffWordsProd} in $A$.
This requires $O(3^n)$ operations corresponding to sets 
$J \subseteq I \subseteq W$ that occur when to the coefficient of $X^I$
a term is added as result of multiplication with $(1 + \langle \beta_J\rangle \cdot \Gamma(|J|)\cdot X^J)$.
\QEDA
\CR
(This is better than using \eqref{eq:SimpleFormulaA} directly, since the number of different set partitions 
grows like $\left(\frac{n}{e\cdot \log n}\right)^n$ - see e.g. \cite{deBruijn}, chapter 6.2; and it is also better than 
using \eqref{eq:product_power_alpha}, which would have resulted in $O(m\cdot 3^n)$ steps.) 

\section{Proof of Lemma \ref{Hausdorff}} 
\begin{namedthm}{Lemma \ref{Hausdorff}}
Let $n, \alpha$ be fixed; then for any $m\geq 2^n$ and $\beta(w_i|z_j)$ given for $j\leq m-1$, the 
function $\BR^n\arrow\BR$ given by 
\[
    \beta(w_1|z_m),...,\beta(w_n|z_m) \mapsto \tilde p(w_1,...,w_n|\alpha, \beta)
\]
is a polynomial of degree $\leq (1,1,...,1)$.\CR
Varying the inputs $\beta(w_i|z_j)$ given for $j\leq m-1$ we obtain
a subset of the $2^n$-dimensional vector space of all possible polynomials, 
this subset has Hausdorff dimension $2^n-1$.
\end{namedthm}
\noindent
{\bf Proof:}
If we abbreviate $\beta_{\{i\}}(z)$\ to $\beta_i$, the Taylor series \eqref{eq:Pochhammer} gives for one factor in \eqref{eq:product_power_alpha}\ the series
 \begin{eqnarray}
    \Big(1-\sum_{i=1}^r \beta_i X_i\Big)^{-\alpha}
  &=& 1 +\alpha \sum_{i=1}^r \beta_i X_i 
        +\frac{\alpha(\alpha+1)}{2!}\Big(\sum_{i=1}^r \beta_i X_i \Big)^2
        +... \nonumber\\
  &\equiv& \sum_{I\subseteq W} (\alpha)_{|I|} \cdot \beta_I(z)\cdot X^I \ \mod (X_1^2,...,X_r^2)
   \label{eq:PowerTaylor}
\end{eqnarray}
Therefore the coefficients of the polynomial
of this Lemma are apart from the constant factors $(\alpha(z_{m}))_{|I\setminus J|}$
just the $\tilde p(J)$ with $J \subseteq W$, computed with all causes except the last.
So we
want show that the subset of all possible coefficient vectors
$\tilde p(J)$ in $\BR^{2^n}$\ is $2^n-1$ - dimensional.
To compute the dimension of the possible vectors of $\tilde{p}(J)$, we first note that this is the same as the dimension of the possible vectors 
$\langle \beta_J \rangle$: In the expression
\[
\tilde{p}(J) = \sum_{\pi \in \calP art(J)} 
         \prod_{K \in \pi} 
         \Gamma(|K|) \cdot \langle \beta_K \rangle
\]
we have one term corresponding to the coarsest partition $\pi = \{J\}$ which is $\langle \beta_J \rangle$, and all other terms involve only smaller sets $K$.
Therefore we can compute $\langle \beta_J \rangle$ from $S(J)$ and the $\langle \beta_K \rangle$ for $|K|<|J|$. So by going through the $J$ in an order of nondecreasing $|J|$ we can determine all the $\langle \beta_J \rangle$ from the $\tilde{p}(J)$.\CR
These are  $2^n$ numbers, but one does not contain information about the $\beta_i$: We have $\langle \beta_{\emptyset} \rangle = \sum_{z} \alpha(z)$, which we assume to be fixed.\CR
To show that none of the $2^n-1$ numbers $\langle \beta_J \rangle$ for $J\neq \emptyset$ are determined by the others, we show that the image of the map $\mu:\BR^{n \cdot (m-1)} \rightarrow \BR^{(2^n-1)}$ given by 
\begin{eqnarray*}
 \big(\beta_1(z_1),...,\beta_n(z_{m-1})\big)&\mapsto& 
  \big(\langle \beta_{\{1\}} \rangle, ...
     , \langle \beta_W \rangle \big) 
\end{eqnarray*}
contains a neighborhood of a point. This means that at least near this point any combination of the ``mixed moments''
$\langle \beta_J\rangle$ can occur, i.e. none is a function of the others.
\CR
To show this, it is sufficient to find a point $\vec{\beta} \in \BR^{n \cdot (m-1)}$ which describes possible input probabilities $\beta(w_i|z_j)$ such that the differential of $\mu$ at $\vec{\beta}$ is of full rank, and for that we will exhibit a map $\psi:\BR^{2^n-1}\rightarrow \BR^{n \cdot( m-1)}$ such that the differential of $\mu \circ \psi:\BR^{2^n-1}\rightarrow \BR^{2^n-1}$ has full rank.\CR
We label the coordinates of $\BR^{2^n-1}$ as $x_J$ for sets 
$\emptyset\neq J \subseteq I$. Define $\psi$ to map the vector $\vec{x}$ with coordinates $x_J$ to the functions $\beta_i$ for $i=1,2,...,n$ on a set of $k = 2^n-1 \leq m-1$ points $z_k$, which we again label by sets $\emptyset \neq J \subseteq I$, such that
\[
  \beta_i(z_J) = \begin{cases}
       x_J  & \hbox{for}\ i\in J  \\
       0    & \hbox{otherwise.}
   \end{cases}
\]
For these functions we have for $\emptyset \neq K \subseteq I$:
\[
   \langle \beta_K \rangle 
       = \sum_{\emptyset \neq J \subseteq I} \beta_K(J)
       = \sum_{K\subseteq J \subseteq I}  x_J^{|K|}
\]
Since $\langle p_K \rangle$ is the ``$K$'' coordinate of $\mu\circ\psi(\vec{x})$,
the differential is given by the matrix with entries
\[
   \frac{\partial}{\partial x_J} \langle \beta_K\rangle =
   \begin{cases}
       |K| \cdot x_J^{|K|-1}  & \hbox{for}\ K \subseteq J  \\
       0    & \hbox{otherwise.}
   \end{cases}
\]
If we order the index sets in an order of non-decreasing cardinality, this is a triangular matrix, and if we choose a point given by small positive coordinates $x_J$, all entries on the diagonal are positive, so this square matrix has positive determinant and hence full rank, and the $\beta(w_i|z_j)$\ are small enough that $\sum_{i=1}^m \beta(w_i|z_j) <\ 1$ so that the point $\psi(\vec x)$ lies in the domain of definition of our map
\[
    \beta(w_1|z_m),...,\beta(w_n|z_m) \mapsto \tilde p(w_1,...,w_n|\alpha, \beta)
\]
\QEDA

\section{Proof of Theorem 3}
\begin{namedthm}{Theorem 3}
If there is a polynomial time algorithm to compute exactly the $p(w_1,...,w_n|\alpha,\beta)$,
there is also a polynomial time algorithm to compute exactly the permanent of a 0-1 matrix, in particular this
would imply P=NP.
\end{namedthm}
\noindent
{\bf Proof:}
If there was such an algorithm, we could apply it in particular to the case that $n=m$ and all $\alpha(z)$ are equal. Thus we would have a polynomial time algorithm that takes as input a $n\times n$ matrix $\beta(\tilde{w}_i | z_j)$ and a number $\alpha$ and compute $\tilde{p}(W)$, which is a polynomial in these input data.
This polynomial is defined even for values for which we have no statistical interpretation, in particular for $\alpha = -1$.
For this value of $\alpha$ we get from 
\eqref{eq:product_power_alpha} that $\tilde{p}(W)$ is just $(-1)^n$ times the sum of all products $\beta(\tilde{w}_i|z_{\sigma(i)})$ for all possible permutations $\sigma:\{1,..,n\} \rightarrow \{1,...,n\}$. But 
this is (apart from the factor $(-1)^n$) just the permanent of the matrix
$\beta(\tilde{w}_i | z_j)$. Now a theorem of Valiant says that if there is a polynomial time algorithm to compute the permanent of a $0,1$-matrix,
we would have $\#P = FP$ and hence also $NP = P$ (see e.g. Theorem 17.11 in \cite{AroraBarak}).

\section{Proof of Theorem 4}
\begin{namedthm}{Theorem 4}
If the interaction graph has tree width $w$, we can compute $p(w_1,...,w_n|\alpha,\beta)$ in time 
$O(n\cdot 3^w + m\cdot n\cdot 2^w)$.
\end{namedthm}
\noindent
{\bf Proof:}\CR
Recall the notion of a tree decomposition of a graph (e.g. \cite{Diestel}, chapter 12.4):\CR
A tree decomposition of $(V,E)$ is a tree $T$ in which all vertices $t\in T$ are labeled by subsets $V_t$ of $V$ such that
\begin{enumerate}
  \item $V$ is the union of the $V_t$
  \item Each clique of $V$ lies in some $V_t$
  \item If $t_2\in T$ lies on a path between $t_1$ and $t_3$, we have
        $ V_{t_1} \cap V_{t_3} \subseteq  V_{t_2}$.
\end{enumerate}
The width of such a tree decomposition is the maximum of the $|V_t|-1$, 
and the tree width of a graph is the minimum width of a tree decomposition. One can show that every graph with $n$ nodes and tree width $w$ has a tree decomposition with width $w$ which also has at most $n$ nodes in the tree.\CR
For example this graph:\CR
\includegraphics[width=0.7\textwidth]{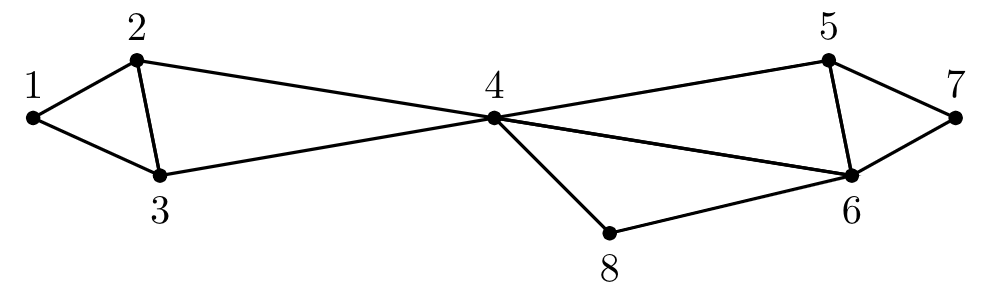}
\CR
has as a tree decomposition:\CR
\includegraphics[width=0.68\textwidth]{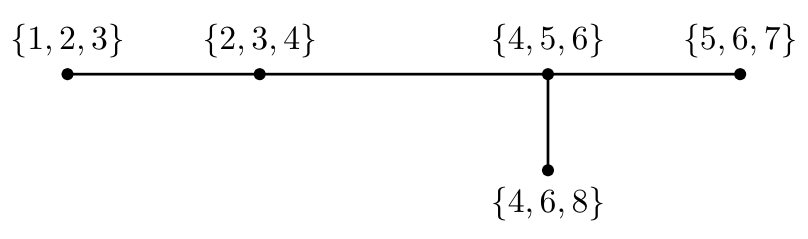}
\CR
and hence the tree width of the above graph is at most 2.\CR
We will assume we are given such a tree decomposition of width $w$; we select one node of the tree $T$ as root and orient every edge such that it points towards this node:\CR
\includegraphics[width=0.71\textwidth]{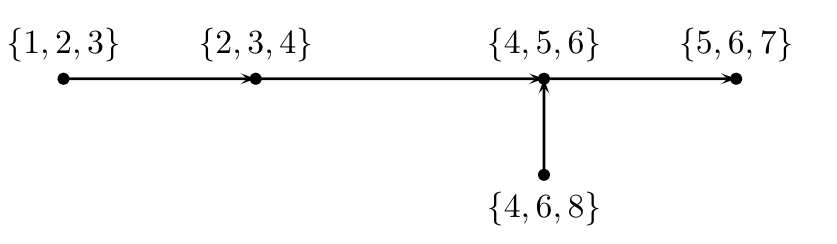}
\CR
The second condition in the definition of a tree decomposition allows us to select a function $\phi$ that assigns to each subset $J \subseteq W$ with $\langle \beta_J \rangle > 0$ a node $\phi(J)\in T$ such that $J\subseteq V_{\phi(J)}$. With this notation the number $\tilde p(w_1,...,w_n)$ is the coefficient of $X_1\cdot...\cdot X_r$ in the product
\[
   S(X) \equiv  \prod_{t\in T}\quad \prod_{J\subseteq \phi^{-1}(t)} 
    \Big(1 + \langle \beta_J\rangle \Gamma(|J|)\cdot X^J\Big) 
    \mod \big(X_1^2,...,X_r^2\big)
\]
We will see that we can compute this using only algebras of dimension $\leq 2^{w+1}$.
To do so, we start with computing 
\[
    \prod_{J\subseteq \phi^{-1}(t)} 
    \Big(1 + \langle \beta_J\rangle \Gamma(|J|)\cdot X^J\Big) \mod \big(X_1^2,...,X_r^2\big)
\]
for each node $t\in T$. This can be done in the algebra generated by the $X_j$ with $j\in V_t$, which has degree $2^{|V_t|} \leq 2^{w+1}$, the time needed for this is bounded by $O(n\cdot 3^w)$. To find the coefficient of $X_1\cdots X_r$ in the product of these elements, we do a topological sort of the nodes in our directed tree, and then we will go through the $t\in T$ in this order, eliminating a leaf node of the tree in each step. A key property we will be using in this procedure follows from the third condition of ``tree decompositions'': If we ``lose'' a word $i$ by going through an edge of the tree, we will not add it back later on the path to the root. So if we are interested in the coefficient of $X_1\cdots X_r$, we will only need to know the coefficients of monomials which already contain the variable $X_i$.
This means we can use the following procedure for each node $t\in T$:\CR
\begin{itemize}
  \item If $t$ is the root node, we are done - output the coefficient of $X^{V_t}$.
  \item Otherwise $t$ is a leaf node and there is exactly one edge $t\rightarrow s$. 
    From the polynomial at $t$, delete all monomials that do not contain all $X_i$ for $i\in V_t \backslash V_s$ (the variables that we ``lose''), and divide the result by the product of these $X_i$, this is now a polynomial in the variables $X_j$ with $j\in V_t\cap V_s$. Now multiply this with the polynomial at $s$, store the result at $s$, and delete node $t$.
\end{itemize}
This procedure can again be done in $O(n\cdot 3^w)$ steps. Adding $O(m\cdot n\cdot 2^w)$ for computing the $\langle \beta_J \rangle$ with $J\subseteq V_t$ gives complexity $O(n\cdot 3^w + m\cdot n\cdot 2^w)$, which is better than the previous $O(3^n + m\cdot 2^n)$ when the tree width is significantly (at least by $\log_2(n)$) smaller than $n$.

\section{Proof of Lemma \ref{mixtureFormula}}
\begin{namedthm}{Lemma \ref{mixtureFormula}}
\begin{equation}
   \BE[\theta_z] = \frac{\alpha(z)}{n+|\alpha|} 
       \sum_{J \subseteq W} \beta_J(z) \cdot |J|! \cdot \frac{\tilde p(W \setminus J)}{\tilde p(W)} 
   \label{eq:BayesianEstimateA}
\end{equation}
\end{namedthm}
\noindent
{\bf Proof:}\CR
For $J\in\calM(W)$ write $J^+$ for the multiset $J + \{w_z\}$.  From the product decomposition for observations \eqref{eq:AllDiffWordsProd} we get for the generating function in $r+1$ variables (we denote the new variable for our virtual observation by $X^\prime$):
\begin{eqnarray*}
  S(X,X^\prime)
    &\equiv& 
        S(X) \cdot \prod_{J\in\calM(W)} \exp\left(\alpha(z) \beta_J(z)\cdot |J| 
        \cdot \frac{X^J\cdot X^\prime}{J!}\right) \\
    &\equiv& S(X) \cdot \left(1 + \alpha(z) \sum_{J\in\calM(W)} \beta_J(z)\cdot |J|!\cdot
            \frac{X^J\cdot X^\prime}{J!}  \right) 
         \quad \mod X^{\prime2}
\end{eqnarray*}
On the other hand this is by definition
\[
   S(X,X^\prime) \equiv S(X) + \sum_{J\in \calM(W)} \tilde p(J^+) \frac{X^J\cdot X^\prime}{J!}
   \mod X^{\prime2}
\]
Equating the coefficient of $X^I\cdot X^\prime$ gives
\begin{eqnarray}
   \BE[\theta_z] &=& \frac{p(I^+)}{p(I)}  
        = \frac{1}{n+|\alpha|}\cdot \frac{\tilde p(I^+)}{\tilde p(I)}
        \nonumber \\
     &=& \frac{\alpha(z)}{n+|\alpha|} 
       \sum_{\emptyset \leq J \leq I} \binom{I}{J} \beta_J(z) 
       \cdot |J|! \cdot \frac{\tilde p(I-J)}{\tilde p(I)} \nonumber\\
\end{eqnarray}
The Lemma is now the special case of $I=W$.

\section{A lemma about rational functions}
The proofs of Theorems 2' and 3' rely on the fact that under some conditions we can 
reconstruct the coefficients of a rational function from the function values.
We state and prove here the purely algebraic facts that we will use.
\begin{lemma}\thlabel{rationalAlg}
\CR
\begin{enumerate}
  \item[a)] 
  Given a rational function 
  \[
     f(X) = \frac{1 + c_1 X + c_2 X^2 + ... + c_n X^n}{1 + d_1 X + d_2 X^2 + ... + d_n X^n}
  \]
  with $c_1,..,c_n,d_1,...,d_n\in\BR$, we have
  \begin{eqnarray*}
   \lefteqn{\frac{\partial^i}{\partial X^i} f(X)|_{X=0}}  \\
     &=& i! \cdot (c_i -d_i) + \hbox{polynomial in other}\ c_k, d_k, k<i
  \end{eqnarray*}
  \item[b)]Given a rational function 
    \[
       f(X_1,...,X_n) = \frac{\sum_{K\subseteq I} c_K\cdot X^K}
                                  {\sum_{K\subseteq I} d_K\cdot X^K}
    \]
    with $I = \{1,...,n\}$ and constant coefficients $c_\emptyset = d_\emptyset = 1$, we have 
    \begin{eqnarray*}
       \lefteqn{ \frac{\partial}{\partial X_1}\frac{\partial}{\partial X_2}\cdots 
          \frac{\partial}{\partial X_n} f|_{(0,0,...,0)} }\\
        &=& c_I - d_I +    \hbox{polynomial in other}\ c_K,d_K, \ K\subset I
    \end{eqnarray*}
\end{enumerate}
\end{lemma}
\noindent
{\bf Proof:}\CR
We can see this basically ``without calculation'' if we look at the structure of the involved formulas.\CR
First note that the denominator in both cases has constant term 1, so we can compute the quotient as 
formal power series and evaluate the derivatives also in that domain. Furthermore, we can treat the
coefficients $c_k,d_k$ or $c_K,d_K$ as symbols, so we are computing the quotient in the ring
\[
   R := \BR[c_1,...,c_n,d_1,...,d_n][[X]]
\]
in the first case and a similar formal power ring in $X_1,...,X_n$ in the second case.
We treat first case a.
\CR
For the division we use the formula 
\[
   \frac{Y}{1+Z} = Y - YZ + YZ^2 -+...
\]
by substituting
\begin{eqnarray*}
  Y &=& 1 + c_1 X + c_2 X^2 + ... + c_n X^n \\
  Z &=&  d_1 X + d_2 X^2 + ... + d_n X^n
\end{eqnarray*}
This gives a power series
\[
   1 + P_1\cdot X + P_2\cdot X^2 + ...
\]
with certain polynomials $P_1,P_2,...\in \BR[c_1,...,c_n, d_1,...,d_n]$. Furthermore, note that in $Y$ and $Z$ 
each term $c_j$ or $d_j$ is ``balanced'' by a $X^j$, and this is preserved when we add and multiply such terms.
So for all monomials (products of $c_j$ and $d_j$) in the polynomials $P_k$ the indices of the $c$ and $d$ have to add up to $k$.
In particular, each $P_k$ can only involve $c_j$ and $d_j$ for $j\leq k$.\CR
Since we are not interested in the exact contributions of $c_k, d_k$ with
$k<i$, and also not in the coefficients of $X^k$ for $k>i$, we can look at this calculation modulo the ideal 
$(c_1,..,c_{i-1},d_1,...,d_{i-1}, X^{i+1})$. Then the calculation 
becomes
\[
   f(X) \equiv \frac{1+c_i\cdot X^i}{1+d_i\cdot X^i} \equiv 1 + (c_i-d_i)\cdot X^i
\]
from which we see that
\[
    P_i\equiv c_i-d_i \mod (c_1,..., c_{i-1},d_1,...,d_{i-1})
\]
Since the indices of a product of the $c_k, d_k$ occuring in the polynomial $P_i$ must add up to $i$, all other
monomials can only contain $c_k, d_k$ for $k<i$, from which we get 
\[
   P_i = c_i - d_i + \ \hbox{polynomial in}\ c_k,d_k\ \hbox{with}\ k<i
\]
from which the assertion follows.
\smallspace
The case b) follows in the same way by using the ideal generated by all $c_K, d_K$ with $K\subset I$ and 
$X^K$ for all $K\supset I$, and noting that the multi-indices of products of $c_K, d_K$ in $P_I$ have to add up to $I$.
\QEDA
\section{Proof of Theorem 2'}
\begin{namedthm}{Theorem 2'}
Any algorithm that computes $\BE[\theta_z|w_1,...,w_n]$ exactly
\begin{itemize}
  \item using Lipschitz continuous functions and finitely many \textbf{if} statements
  \item going once through the causes $z_j$ and reading in the $\alpha(z_j), \beta(w_i,z_j)$ in order of ascending $j$
  \item and outputs after reading the data of cause $z_j$ what $\BE[\theta_z|w_1,...,w_n]$ would be if this was the
    last cause,
\end{itemize}
needs space $O(2^n)$ and time $O(m\cdot 2^n)$.
\end{namedthm}
\noindent
{\bf Proof:}\CR
At the first look the arguments of the proof of Theorem 2 seem to apply directly to equation \eqref{eq:BayesianEstimateA}:
Again this seems to be a polynomial in the $\beta_j(z)$ which determines its coefficients 
$\tilde p(I\backslash J)/\tilde p(I)$. Since $\tilde p(\emptyset)$ is a constant, these numbers determine all $\tilde p(J)$, and we know already that the possible vectors for them form a $2^n-1$ - dimensional set. However, the $\tilde p(J)$ occurring here already contain the contribution from the last cause $z$, and our argument only works when we only have contributions from the previous causes. To fix this, let $\tilde p(J)$, $S(X)$, etc. be the quantities that incorporate all causes except the last one; we denote the last cause by $z^\prime$, and the quantities that incorporate also $z^\prime$ by $\tilde p^+(J)$, $S^+(X)$. As above we introduce an additional ``virtual'' observation $w^\prime$ and corresponding formal variable $X^\prime$.
Then \eqref{eq:product_power_alpha} gives
\begin{eqnarray*}
  S^+(X,X^\prime)
    &=& S(X) \cdot 
     \Big(1 - \big(\sum_{w\in W} \beta(w|z^\prime)X_w\big)-X^\prime\Big)^{-\alpha(z^\prime)}
\end{eqnarray*}
Using the first two terms in the Taylor series
\begin{eqnarray*}
   (c-X^\prime)^{-\alpha} 
      &=&      c^{-\alpha} (1-X^\prime/c)^{-\alpha} \\
     &\equiv&  c^{-\alpha} (1+ \alpha X^\prime/c) \mod X^{\prime 2}
\end{eqnarray*}
this gives with \eqref{eq:PowerTaylor}
\begin{eqnarray*}
  S^+(X,X^\prime)
    &\equiv&
    S(X) \cdot 
     \Big(1 - \sum_{w\in W} \beta(w|z^\prime)X_w\Big)^{-\alpha(z^\prime)} \\
     & &\quad\cdot
     \Big( 1 + X^\prime \cdot \alpha(z^\prime)\cdot \big(\sum_{w\in W} \beta(w|z^\prime)X_w\big)^{-1}\Big)\\
    &\equiv&
    S(X) \cdot \Big(\sum_{I\subseteq W} \big(\alpha(z^\prime)\big)_{|I|} 
        \beta_I(z^\prime)\cdot X^I\Big)  \\
     & &\ \ +\  X^\prime \cdot \alpha(z^\prime)\cdot S(X) \cdot 
        \Big(\sum_{I\subseteq W} \big(\alpha(z^\prime)+1\big)_{|I|} 
        \beta_I(z^\prime)\cdot X^I\Big)
\end{eqnarray*}
from which we get as a variant of \eqref{eq:BayesianEstimateA}
\begin{equation}
   \BE[\theta_{z^\prime}] = \frac{\alpha(z^\prime)}{n+|\alpha|}\cdot 
    \frac{\sum_{I\subseteq W} (\alpha(z^\prime)+1)_{|I|}
           \cdot \beta_I(z^\prime)\cdot \tilde p(W\backslash I)}
         {\sum_{I\subseteq W} (\alpha(z^\prime))_{|I|}
           \cdot \beta_I(z^\prime)\cdot \tilde p(W\backslash I)}
   \label{eq:ExpNewTopic}
\end{equation}
which is a rational function in the $\beta_i(z^\prime)$. To extract the information about the $\tilde p(J)/\tilde p(W)$ from 
this function we can use part b) of \thref{rationalAlg}; applying this to a sequence of the subsets $I$ of $\{1,...,n\}$ sorted 
in nondecreasing order of $|I|$ and to the rational function $\frac{n+|\alpha|}{\alpha(z^\prime)}\BE[\theta_{z^\prime}]$
and dividing denominator and numerator by $\tilde p(W)$ this gives us the numbers
\[
    \Big((\alpha(z^\prime)+1)_{|I|} -  (\alpha(z^\prime))_{|I|}\Big)\cdot
     \frac{\tilde p(W\backslash I)}{\tilde p(W)}
    \ =\  |I|\cdot (\alpha(z^\prime)+1)_{|I|-1} \cdot \frac{\tilde p(W\backslash I)}{\tilde p(W)}
\]
which finally allows us to conclude like above  that we need all the numbers $\tilde p(J)$, and any ``online'' algorithm to compute $\BE(\theta_z)$  needs to update $2^n-1$ numbers for each cause, so the complexity has to be at least $O(m\cdot 2^n)$ in this case as well.

\section{Proof of Theorem 3'}
\begin{namedthm}{Theorem 3'}
If there is a polynomial time algorithm to compute exactly the $\BE[\theta_z|w_1,...,w_n]$,
there is also a polynomial time algorithm to compute exactly the permanent of a 0-1 matrix, in particular this
would imply P=NP.
\end{namedthm}
We add a new virtual cause $z'$ with $\alpha(z')=1$ and $\beta(w|z')=\epsilon$ for all $w$. Then
$\BE[\theta(z')|w_1,...,w_n]$ as a function of $\epsilon$ is given by \eqref{eq:ExpNewTopic}, for our $\alpha,\beta$
this simplifies to 
\begin{equation}
   \BE[\theta_{z^\prime}] = \frac{1}{n+|\alpha|}\cdot 
    \frac{\sum_{I\subseteq W} (|I|+1)!
           \cdot \epsilon^{|I|}\cdot \tilde p(W\backslash I)}
         {\sum_{I\subseteq W} |I|!
           \cdot \epsilon^{|I|}\cdot \tilde p(W\backslash I)}
   \label{eq:ExpNewTopic2}
\end{equation}
where we use the convention that $0!=1$.\CR
Introduce for $k=0,...,n$ the numbers
\[
    d_k := \sum_{I\subseteq W, |I|=k} k! \cdot \tilde p(W\backslash I) / \tilde p(W)
    \quad \hbox{and}\quad
    c_k := (k+1)\cdot d_k
\]
Then $c_0 = d_0 = 1$, $d_n = n! / \tilde p(W)$ and 
\[
   f(\eps) := (n+|\alpha|)\cdot \BE[\theta_{z^\prime}] 
       = \frac{1 + c_1 \eps + c_2 \eps^2 + ... c_n \cdot \eps^n}{1 + d_1 \eps + d_2 \eps^2 + ... d_n \cdot \eps^n}
\]
So the question becomes if we can use values of $f(\eps)$ to compute $d_n$.
\smallspace
If we compute $f(\eps)$ at $2n+1$ different points $\eps_1,...,\eps_{2n+1}$, then the rational function
$P(\eps)/Q(\eps)$ of $\eps$ with degree of $P$ and $Q$ at most $n$ is uniquely determined by the 
condition $f(\eps_i)=P(\eps_i)/Q(\eps_i)$, and it can be found by determining a nontrivial solution
of the $2n+1$ linear equations $f(\eps_i)\cdot Q(\eps_i)=P(\eps_i)$ in the $2n+2$ coefficients of $P$ and $Q$.
(see e.g. Theorem 5.9, p.132 of \cite{Rivlin}).\CR
While this allows us to compute a representation of the rational function $f(\eps)$, it does not guarantee the uniqueness
of $P$ and $Q$ (they could have a common factor of degree $\geq 1$). However, we get uniqueness from the
additional condition $c_k = (k+1) \cdot d_k$: Part a) of \thref{rationalAlg} allows us to compute the unique 
$d_1, d_2,...,d_n$ from a representation of $f(\eps)$
as $P(\eps)/Q(\eps)$, which in turn gives $\tilde p(w_1,...,w_n|\alpha,\beta) = \tilde p(W) = n! / d_n$
which reduces Theorem 3' to Theorem 3.

\begin{thebibliography}{}
\setlength{\itemindent}{-\leftmargin}
\makeatletter\def\@biblabel#1{\def\citename##1{##1}[#1]\hfill}\makeatother

\bibitem{Hofmann}
Thomas Hofmann (1999).
\newblock Probabilistic Latent Semantic Analysis.
\textit{Proceedings of Uncertainty in Artificial Intelligence, UAI'99 Stockholm}, 289-296

\bibitem{BleiNgJordan}
David M. Blei, Andrew Y. Ng, Michael I. Jordan (2003).
\newblock Latent Dirichlet Allocation.
\textit{Journal of Machine Learning Research 3}, 993-1022.

\bibitem{Pritchard}
Jonathan K. Pritchard, Matthew Stephens and Peter Donnelly.
\newblock Inference of Population Structure Using Multilocus Genotype Data.
\textit{GENETICS June 1, 2000 vol. 155 no. 2}, 945-959

\bibitem{Dickey1}
J. Dickey (1983).
\newblock Multiple hypergeometric functions: Probabilistic interpretations and statistical uses.
\textit{Journal of the American Statistical Association, 78}, 628-637

\bibitem{SontagRoy}
David Sontag, Daniel M. Roy (2011).
\newblock Complexity of Inference in Latent Dirichlet Allocation.
\textit{NIPS 2011}

\bibitem{Falconer}
Kenneth Falconer (2003).
\newblock \textit{Fractal Geometry, Mathematical Foundations and Applications}.
2nd edition, Wiley

\bibitem{AroraBarak}
Sanjeev Arora, Boaz Barak (2009).
\newblock \textit{Computational Complexity, A Modern Approach}.
\newblock Cambridge University Press

\bibitem{Diestel}
Reinhard Diestel (2017).
\newblock \textit{Graph Theory}. 5th edition, Springer

\bibitem{deBruijn}
N. G. de Bruijn (1958).
\newblock \textit{Asymptotic Methods in Analysis}. North Holland

\bibitem{Wasserman}
Larry Wasserman (2004).
\newblock \textit{All of Statistics}. Springer

\bibitem{GhoshRamamoorthi}
J.K. Ghosh, R.V. Ramamoorthi (2003)
\newblock \textit{Bayesian Nonparametrics}. Springer

\bibitem{Rivlin}
Theodore J. Rivlin (1969)
\newblock \textit{An introduction to the approximation of functions}. Blaisdell
\end{thebibliography}
\end{document}